\documentclass[final]{cvpr}

\usepackage{times}
\usepackage{epsfig}
\usepackage{graphicx}
\usepackage{amsmath}
\usepackage{amssymb}

\usepackage{amsmath,amsfonts,bm}

\def\eqref#1{equation~\ref{#1}}

\def\1{\bm{1}}

\def\vzero{{\bm{0}}}
\def\vone{{\bm{1}}}
\def\vmu{{\bm{\mu}}}

\def\vp{{\bm{p}}}
\def\vq{{\bm{q}}}

\def\vs{{\bm{s}}}

\def\vw{{\bm{w}}}
\def\vx{{\bm{x}}}

\def\vz{{\bm{z}}}

\def\mW{{\bm{W}}}

\DeclareMathAlphabet{\mathsfit}{\encodingdefault}{\sfdefault}{m}{sl}
\SetMathAlphabet{\mathsfit}{bold}{\encodingdefault}{\sfdefault}{bx}{n}

\def\gD{{\mathcal{D}}}

\def\gS{{\mathcal{S}}}

\def\sR{{\mathbb{R}}}

\newcommand{\R}{\mathbb{R}}

\usepackage{colortbl}
\usepackage{url}
\usepackage[utf8]{inputenc} 
\usepackage[T1]{fontenc}    
\usepackage{booktabs}       
\usepackage{amsfonts}       
\usepackage{nicefrac}       
\usepackage{microtype}      
\usepackage{xcolor}
\usepackage{bm}
\usepackage{bbm}
\usepackage{amsbsy}
\usepackage{amsmath}
\usepackage{cases}
\usepackage{enumitem}
\usepackage{lipsum}
\usepackage{caption}
\usepackage{graphicx}
\usepackage{subcaption}
\usepackage{multirow}
\usepackage[titletoc]{appendix}
\usepackage{grffile}
\usepackage{diagbox}
\usepackage{pifont}
\usepackage{amssymb}
\usepackage{multirow}
\usepackage{multicol}
\newcolumntype{I}{!{\vrule width 1pt}}
\usepackage{color,colortbl}
\definecolor{lightgray}{HTML}{F0F0F0}  
\definecolor{rowbackground}{HTML}{F9F9F9}
\definecolor{Gray}{gray}{0.9}
\definecolor{LightCyan}{rgb}{0.88,1,1}
\definecolor{myblue}{rgb}{.935,.935,.99}
\usepackage{textpos}

\usepackage[pagebackref=false,breaklinks=true,colorlinks,bookmarks=false]{hyperref}
\def\cvprPaperID{741} 
\def\confYear{CVPR 2021}

\begin{document}
	
	\title{Improving Calibration for Long-Tailed Recognition}
	
	\author{
		Zhisheng Zhong\quad\quad Jiequan Cui \quad\quad Shu Liu \quad\quad Jiaya Jia \vspace{.3em}
		\\
		Chinese University of Hong Kong \quad\quad SmartMore \vspace{.3em}
	}

	\maketitle
	\pagestyle{empty}
	\thispagestyle{empty}

	\begin{textblock*}{.8\textwidth}[.5,0](0.5\textwidth, -0.07\textwidth)
		\centering
		{\small Code: \url{https://github.com/Jia-Research-Lab/MiSLAS}}
	\end{textblock*}
	
	
	\begin{abstract}
		Deep neural networks may perform poorly when training datasets are heavily class-imbalanced. Recently, two-stage methods decouple representation learning and classifier learning to improve performance. But there is still the vital issue of miscalibration. To address it, we design two methods to improve calibration and performance in such scenarios. Motivated by the fact that predicted probability distributions of classes are highly related to the numbers of class instances, we propose label-aware smoothing to deal with different degrees of over-confidence for classes and improve classifier learning. For dataset bias between these two stages due to different samplers, we further propose shifted batch normalization in the decoupling framework. Our proposed methods set new records on multiple popular long-tailed recognition benchmark datasets, including CIFAR-10-LT, CIFAR-100-LT, ImageNet-LT, Places-LT, and iNaturalist 2018.
	\end{abstract}

	\vspace{-10pt}
	\section{Introduction}
	
	\begin{figure}[t]
		\vspace{18pt}
		\centering
		\begin{subfigure}{0.48\linewidth}
			\centering
			{\small{\quad\quad\quad  Org. CIFAR-100}}
		\end{subfigure} 
		\hfill
		\begin{subfigure}{0.48\linewidth}
			\centering
			{\small \ \ \  CIFAR-100-LT, IF100}
		\end{subfigure} \\
		\vspace{3pt}
		\includegraphics[width=0.48\textwidth]{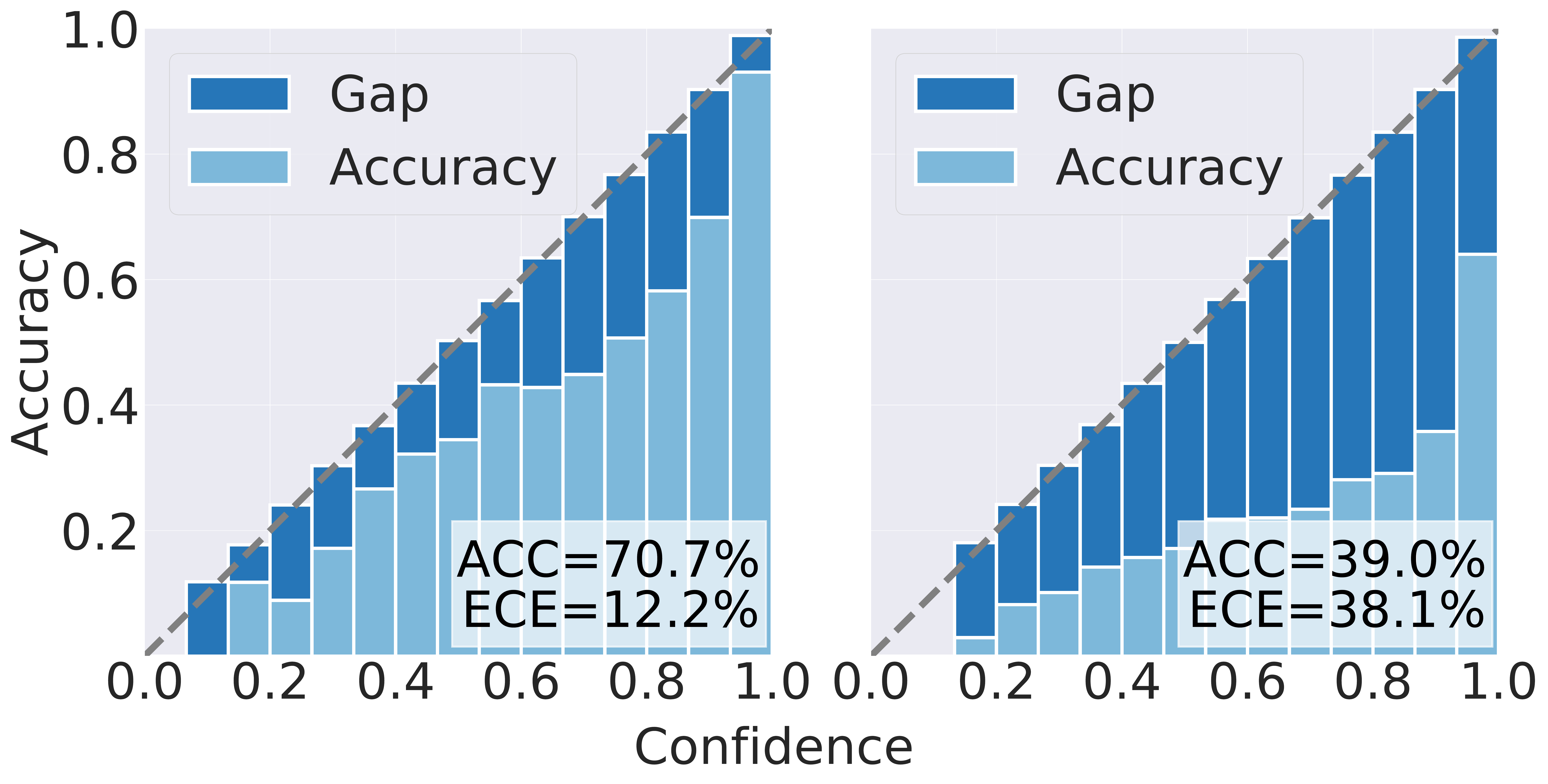} 
		\begin{subfigure}{0.5\linewidth}
			\centering
			{\small{\quad\quad\ CIFAR-100-LT, IF100, cRT}}
		\end{subfigure} 
		\hfill
		\begin{subfigure}{0.48\linewidth}
			\centering
			{\small \ \ CIFAR-100-LT, IF100, LWS}
		\end{subfigure} \\
		\vspace{3pt}
		\includegraphics[width=0.48\textwidth]{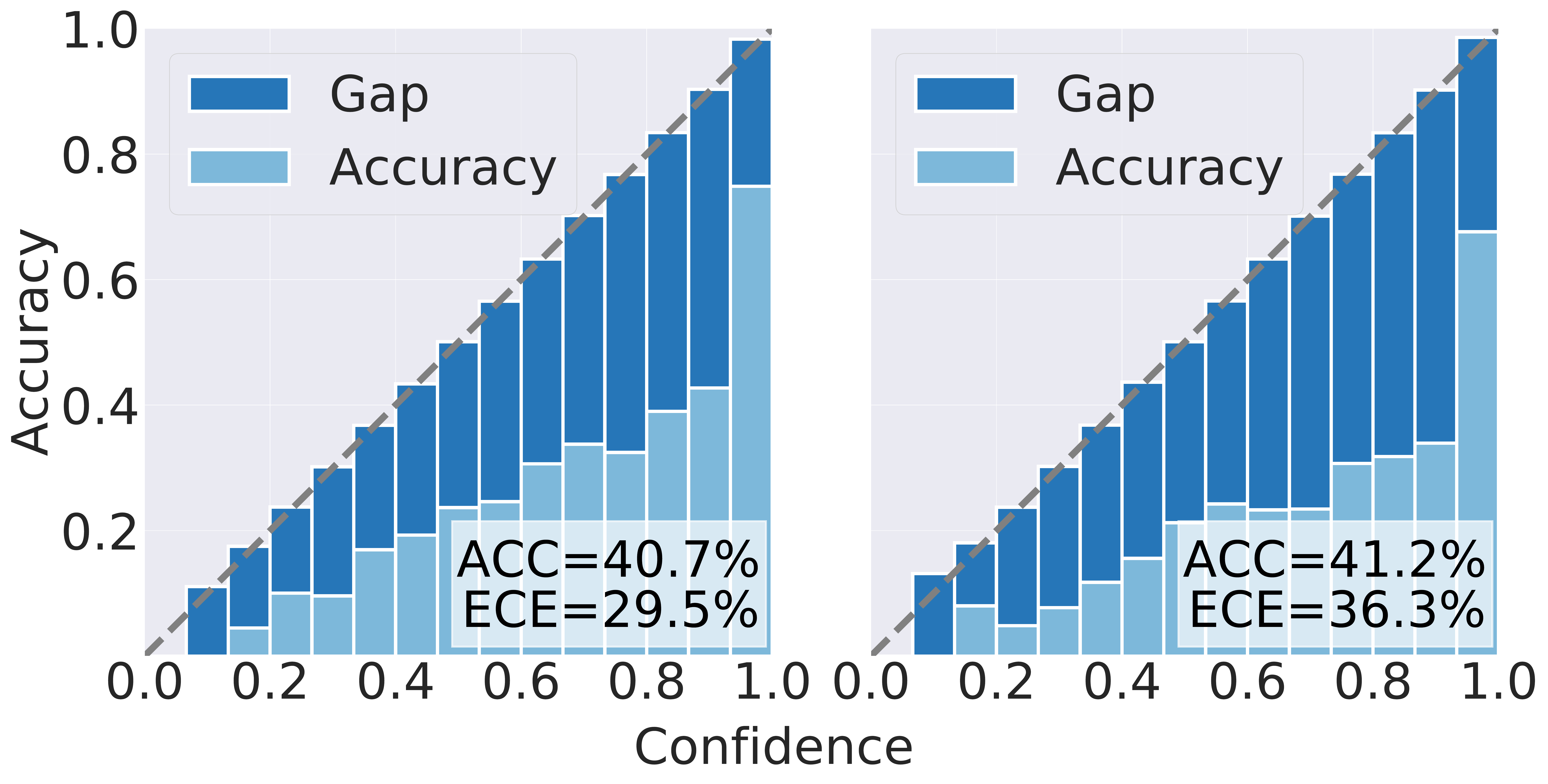} 
		\vspace{-18pt}
		\caption{Reliability diagrams of ResNet-32. From top left to bottom right: the plain model trained on the original balanced CIFAR-100 dataset, the plain model, cRT, and LWS trained on CIFAR-100-LT with IF 100.}
		\vspace{-20pt}
		\label{fig:overcondifence}
	\end{figure}
	
	%
	%
	
	With numerous available large-scale and high-quality datasets, such as ImageNet~\cite{imagenet}, COCO~\cite{coco}, and Places~\cite{places}, deep convolutional neural networks~(CNNs) have made notable breakthrough in various computer vision tasks, such as image recognition~\cite{alexnet, resnet}, object detection~\cite{fasterrcnn}, and semantic segmentation~\cite{cordts2016cityscapes}. These datasets are usually artificially balanced with respect to the number of instances for each object/class. However, in many real-world applications, data may follow unexpected long-tailed distributions, where the numbers of instances for different classes are seriously imbalanced. When training CNNs on these long-tailed datasets, the performance notably degrades. To address this terrible issue, a number of methods were proposed for long-tailed recognition.

	Recently, many two-stage approaches have achieved significant improvement comparing with one-stage methods. Deferred re-sampling~(DRS, \cite{ldam}) and deferred re-weighting~(DRW, \cite{ldam}) first train CNNs in a normal way in Stage-1. DRS tunes CNNs on datasets with class-balanced resampling while DRW tunes CNNs by assigning different weights to classes in Stage-2. Zhou \etal~\cite{bbn} proposed bilateral branch network~(BBN) in one stage to simulate the process of DRS by dynamically combining instance-balanced sampler and the reverse-balanced sampler. Kang \etal~\cite{decouple} proposed two-stage decoupling models, classifier re-training~(cRT) and learnable weight scaling~(LWS), to further boost performance, where decoupling models freeze the backbone and just train the classifier with class-balanced resampling in Stage-2.

	Confidence calibration~\cite{niculescu2005predicting, calibration} is to predict probability by estimating representative of true correctness likelihood. It is important for recognition models in many applications~\cite{bojarski2016end, jiang2012calibrating}. Expected calibration error~(ECE) is widely used in measuring calibration of the network. To compute ECE, all $N$ predictions are first grouped into $B$ interval bins of equal size. ECE is defined as:
	$$\text{ECE} = \sum_{b=1}^{B} \frac{|\gS_b|}{N}\bigg|\text{acc}(\gS_b) - \text{conf}(\gS_b)\bigg|\times 100\%,$$
	where $\gS_b$ is the set of samples whose prediction scores fall into Bin-$b$. $\text{acc}(\cdot)$ and $\text{conf}(\cdot)$ are the accuracy and predicted confidence of $\gS_b$, respectively.
	
	Our study shows, because of the imbalanced composition ratio of each class, networks trained on long-tailed datasets are more miscalibrated and over-confident. We draw the reliability diagrams with 15 bins in Fig.~\ref{fig:overcondifence}, which compares the plain cross-entropy~(CE) model trained on the original CIFAR-100 dataset, the plain CE model, cRT, and LWS trained on CIFAR-100-LT with imbalanced factor (IF) 100. It is noticeable that networks trained on long-tailed datasets usually have higher ECEs. The two-stage models of cRT and LWS suffer from over-confidence as well. Moreover, Figs.~\ref{fig:more_rd_cifar10} and \ref{fig:more_rd_imgnet} (the first two plots) in Appendix~\ref{sec:cb} depict that this phenomenon also commonly exists on other long-tailed datasets, such as CIFAR-10-LT and ImageNet-LT.
	
	
	Another issue is that two-stage decoupling ignores the dataset bias or domain shift~\cite{quionero2009dataset} in the two stages. In details, two-stage models are first trained on the instanced-balanced dataset $\gD_{\rm{I}}$ in Stage-1. Then, models are trained on the class-balanced dataset $\gD_{\rm{C}}$ in Stage-2. Obviously, $\displaystyle P_{\gD_{\rm{I}}}(\vx, y) \not= \displaystyle P_{\gD_{\rm{C}}}(\vx, y)$ and distributions of the dataset by different sampling ways are inconsistent. Motivated by transfer learning \cite{li2018adaptive, wang2019transferable}, we focus on the batch normalization~\cite{bn} layer to deal with the dataset bias problem.
	
	In this work, we propose a \textbf{Mi}xup \textbf{S}hifted \textbf{L}abel-\textbf{A}ware \textbf{S}moothing model~(MiSLAS) to effectively solve above issues. Our key contributions are as follows.
	
	\begin{itemize}[itemsep=-0.2em, leftmargin=1em]
		\item  We discover that models trained on long-tailed datasets are much more miscalibrated and over-confident than those trained on balanced data. Two-stage models suffer from this problem as well. 
		\item We find that mixup can remedy over-confidence and have a positive effect on representation learning but a negative or negligible effect on classifier learning. To further enhance classifier learning and calibration, we propose label-aware smoothing to handle different degrees of over-confidence for  classes. 
		\item It is the first attempt to note the dataset bias or domain shift in two-stage resampling methods for long-tailed recognition. To deal with it in the decoupling framework, we propose shift learning on the batch normalization layer, which can greatly improve performance. 
		\item We extensively validate our MiSLAS on multiple long-tailed recognition benchmark datasets -- experimental results manifest the effectiveness. Our method yields new state-of-the-art.
	\end{itemize}


	\section{Related Work}

	\begin{table*}[t]
		\hspace{1pt}
		\begin{minipage}[t]{0.48\textwidth}
			\begin{center}
				\small
				\setlength{\tabcolsep}{4pt}
				\begin{tabular}{l|cc|ccc}
					\toprule[1.5pt]
					\textbf{Mark} & Stg.-1 & Stg.-2  & ResNet-50 & ResNet-101 & ResNet-152 \\
					\midrule
					CE & \rlap{\raisebox{0.3ex}{\hspace{0.4ex}\scriptsize \ding{56}}}$\square$ &  & 45.7 / 13.7 & 47.3 / 13.7 & 48.7 / 14.5\\
					CE & \rlap{\raisebox{0.3ex}{\hspace{0.4ex}\tiny \ding{52}}}$\square$ &  & 45.5 / 7.98  & 47.7 / 10.1  & 48.3 / 10.2 \\
					\midrule
					\multicolumn{1}{>{\columncolor{myblue}[2pt][212.5pt]}l|}{cRT} & \rlap{\raisebox{0.3ex}{\hspace{0.4ex}\scriptsize \ding{56}}}$\square$ & \rlap{\raisebox{0.3ex}{\hspace{0.4ex}\scriptsize \ding{56}}}$\square$ &50.3 / 8.97 & 51.3 / 9.34 & 52.7 / 9.05 \\
					cRT & \rlap{\raisebox{0.3ex}{\hspace{0.4ex}\scriptsize \ding{56}}}$\square$ & 
					\rlap{\raisebox{0.3ex}{\hspace{0.4ex}\tiny \ding{52}}}$\square$ & 50.2 / 3.32 & 51.3 / 3.38 & 52.8 / 3.60 \\
					\multicolumn{1}{>{\columncolor{myblue}[2pt][212.5pt]}l|}{cRT} & \rlap{\raisebox{0.3ex}{\hspace{0.4ex}\tiny \ding{52}}}$\square$ & \rlap{\raisebox{0.3ex}{\hspace{0.4ex}\scriptsize \ding{56}}}$\square$ & \textbf{51.7} / 5.62 & \textbf{53.1} / 6.86 & \textbf{54.2} / 6.02 \\
					cRT & \rlap{\raisebox{0.3ex}{\hspace{0.4ex}\tiny \ding{52}}}$\square$ & \rlap{\raisebox{0.3ex}{\hspace{0.4ex}\tiny \ding{52}}}$\square$ & 51.6 / \textbf{3.13} & 53.0 / \textbf{2.93} & 54.1 / \textbf{3.37} \\	
					\bottomrule[1.5pt]
				\end{tabular}
			\end{center}
		\end{minipage}
		\hspace{8.2pt}
		\begin{minipage}[t]{0.48\textwidth}
			\begin{center}
				\small
				\setlength{\tabcolsep}{4pt}
				\begin{tabular}{l|cc|ccc}
					\toprule[1.5pt]
					\textbf{Mark} & Stg.-1 & Stg.-2  & ResNet-50 & ResNet-101 & ResNet-152 \\
					\midrule
					CE & \rlap{\raisebox{0.3ex}{\hspace{0.4ex}\scriptsize \ding{56}}}$\square$ &  & 45.7 / 13.7 & 47.3 / 13.7 & 48.7 / 14.5\\
					CE & \rlap{\raisebox{0.3ex}{\hspace{0.4ex}\tiny \ding{52}}}$\square$ &  & 45.5 / 7.98  & 47.7 / 10.1  & 48.3 / 10.2 \\
					\midrule
					\multicolumn{1}{>{\columncolor{myblue}[2pt][212.5pt]}l|}{LWS} & \rlap{\raisebox{0.3ex}{\hspace{0.4ex}\scriptsize \ding{56}}}$\square$ & \rlap{\raisebox{0.3ex}{\hspace{0.4ex}\scriptsize \ding{56}}}$\square$ & 51.2 / 4.89 & 52.3 / 5.10 & 53.8 / 4.48 \\
					LWS & \rlap{\raisebox{0.3ex}{\hspace{0.4ex}\scriptsize \ding{56}}}$\square$ & 
					\rlap{\raisebox{0.3ex}{\hspace{0.4ex}\tiny \ding{52}}}$\square$ & 51.0 / 5.01 & 52.2 / 5.38 & 53.6 / 5.50 \\
					\multicolumn{1}{>{\columncolor{myblue}[2pt][212.5pt]}l|}{LWS} & \rlap{\raisebox{0.3ex}{\hspace{0.4ex}\tiny \ding{52}}}$\square$ & \rlap{\raisebox{0.3ex}{\hspace{0.4ex}\scriptsize \ding{56}}}$\square$ & 52.0 / \textbf{2.23} & \textbf{53.5} / \textbf{2.73} & \textbf{54.6} / \textbf{2.46} \\
					LWS & \rlap{\raisebox{0.3ex}{\hspace{0.4ex}\tiny \ding{52}}}$\square$ & \rlap{\raisebox{0.3ex}{\hspace{0.4ex}\tiny \ding{52}}}$\square$ & 52.0 / 8.04 & 53.3 / 6.97 & 54.4 / 7.74 \\
					\bottomrule[1.5pt]
				\end{tabular}
			\end{center}
		\end{minipage}
		\hfill
		\hspace{3pt}
		\caption{Top-1 accuracy (\%) and ECE (\%) of the plain cross-entropy~(CE) model, and decoupling models of cRT (left) and LWS (right), for ResNet families trained on the ImageNet-LT dataset. We vary the augmentation strategies with (\rlap{\raisebox{0.3ex}{\hspace{0.4ex}\tiny \ding{52}}}$\square$), or without (\rlap{\raisebox{0.3ex}{\hspace{0.4ex}\scriptsize \ding{56}}}$\square$) mixup $\alpha=0.2$, on both of the stages.}
		\label{tab:augmentation}
		\vspace{-1pt}
	\end{table*}
	
	\begin{figure*}[t]
		\centering
		\begin{subfigure}{0.48\linewidth}
			\includegraphics[width=\textwidth]{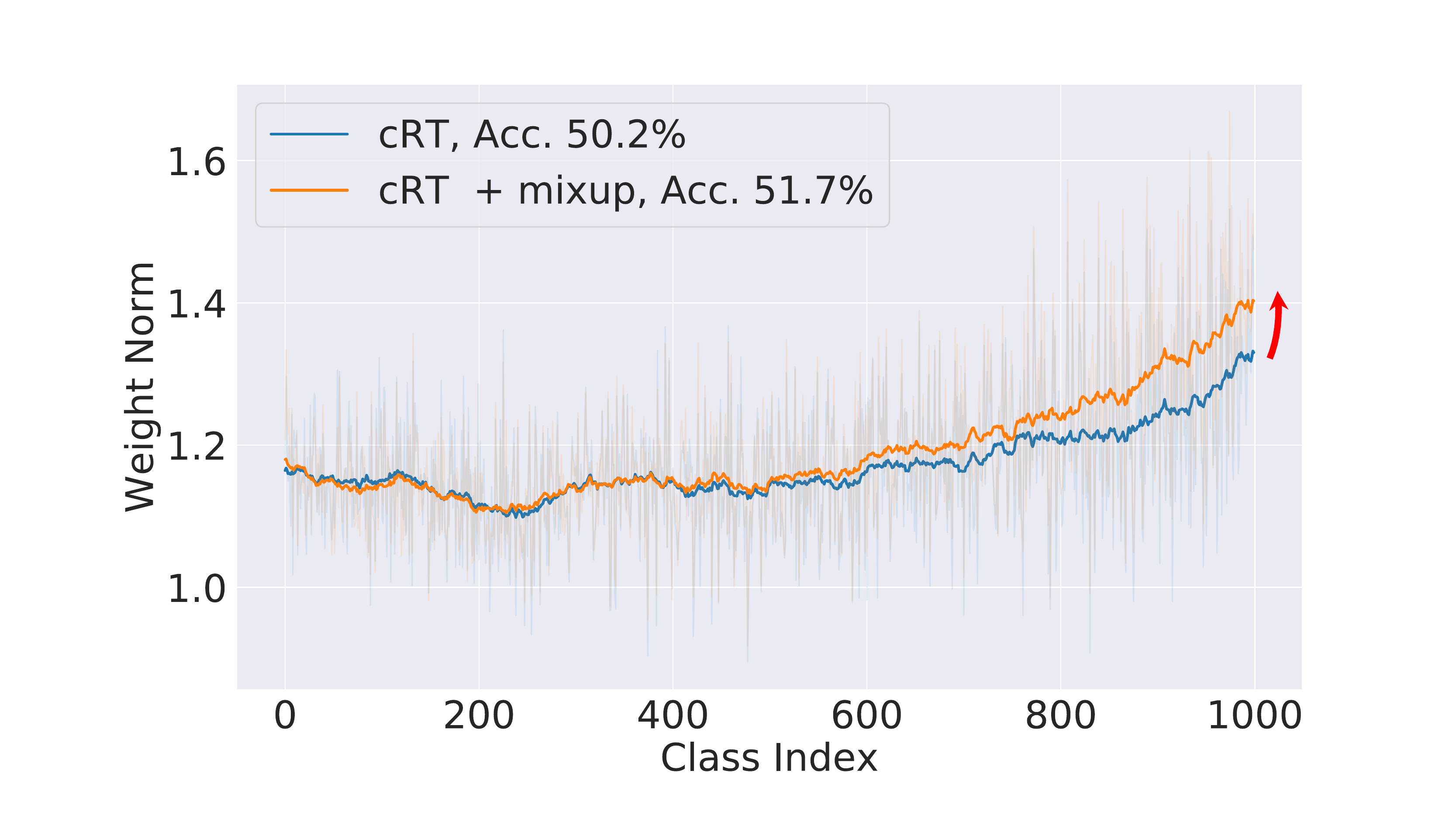}
		\end{subfigure}
		\hspace{12pt}
		\begin{subfigure}{0.48\linewidth}
			\includegraphics[width=\textwidth]{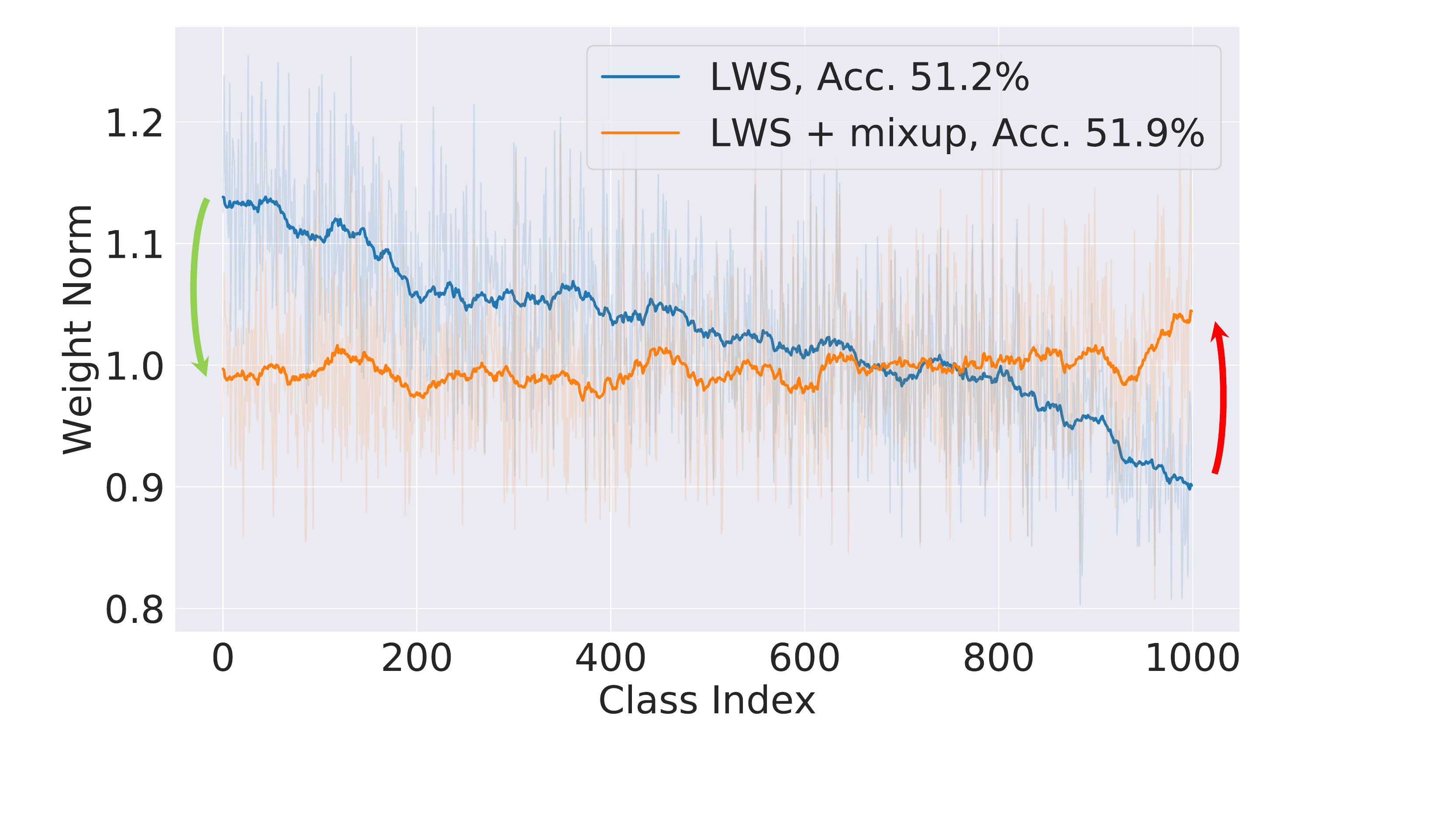}
		\end{subfigure}
		\vspace{-2pt}
		\caption{Classifier weight norms for the ImageNet-LT validation set where classes are sorted by descending values of $N_j$, where $N_j$ denotes the number of training sample for
			Class-$j$. Left: weight norms of cRT with or without mixup. Right: weight norms of LWS with or without mixup. Light shade: true norm. Dark lines: smooth version. \textit{Best viewed on screen}.}
		\label{fig:norm}
		\vspace{-10pt}
	\end{figure*}
	
	\paragraph{Re-sampling and re-weighting.} There are two groups of re-sampling strategies: over-sampling the tail-class images~\cite{shen2016relay, buda2018systematic, byrd2019effect} and under-sampling the head-class images~\cite{japkowicz2002class, buda2018systematic}. Over-sampling is regularly useful on large datasets and may suffer from heavy over-fitting to tail classes especially on small datasets. For under-sampling, it discards a large portion of data, which inevitably causes degradation of the generalization ability of deep models. Re-weighting~\cite{huang2016learning, wang2017learning} is another prominent strategy. It assigns different weights for classes and even instances. The vanilla re-weighting method gives class weights in reverse proportion to the number of samples of classes. 
	
	However, with large-scale data, re-weighting makes deep models difficult to optimize during training. Cui \etal~\cite{effnum} relieved the problem using the effective numbers to calculate the class weights. Another line of work is to adaptively re-weight each instance. For example, focal loss~\cite{lin2017focal, focal_ca} assigned smaller weights for well-classified samples.
	
	\vspace{-10pt}
	\paragraph{Confidence calibration and regularization.} Calibrated confidence is significant for classification models in many applications. Calibration of modern neural networks is first discussed in \cite{calibration}. The authors discovered that model capacity, normalization, and regularization have strong effect on network calibration. mixup~\cite{mixup} is a regularization technique to train with interpolation of input and labels. 
	
	mixup inspires follow-up of manifold mixup~\cite{mmixup}, CutMix~\cite{cutmix}, and Remix~\cite{remix} that have shown significant improvement. Thulasidasan \etal~\cite{mixup_ca} found that CNNs trained with mixup are better calibrated. Label smoothing~\cite{labelsmoothing} is another regularization technique that encourages the model to be less over-confident. Unlike cross-entropy that computes loss upon the ground truth labels, label smoothing computes loss upon a soft version of labels. It relieves over-fitting and increases calibration and reliability~\cite{smooth_ca}.
	
	\vspace{-10pt}
	\paragraph{Two-stage methods.} Cao \etal~\cite{ldam} proposed deferred re-weighting (DRW) and deferred re-sampling (DRS), working better than conventional one-stage methods. Its stage-2, starting from better features, adjusts the decision boundary and locally tunes features. Recently, Kang \etal~\cite{decouple} and Zhou \etal~\cite{bbn} concluded that although class re-balance matters for jointly training representation and classifier, instance-balanced sampling gives more general representations. 
	
	Based on this observation, Kang \etal~\cite{decouple} achieved state-of-the-art results by decomposing representation and classifier learning. It first trains the deep models with instance-balanced sampling, and then fine-tunes the classifier with class-balanced sampling with parameters of representation learning fixed. Similarly, Zhou \etal~\cite{bbn} integrated mixup training into the proposed cumulative learning strategy. It bridges the representation learning and classifier re-balancing. The cumulative learning strategy requires dual samplers of instance-balanced and reversed instance-balanced sampler. 
	
	\vspace{-1pt} 
	
	\section{Main Approach}
	
	\begin{figure*}[t]
		\centering
		\includegraphics[width=0.95\textwidth]{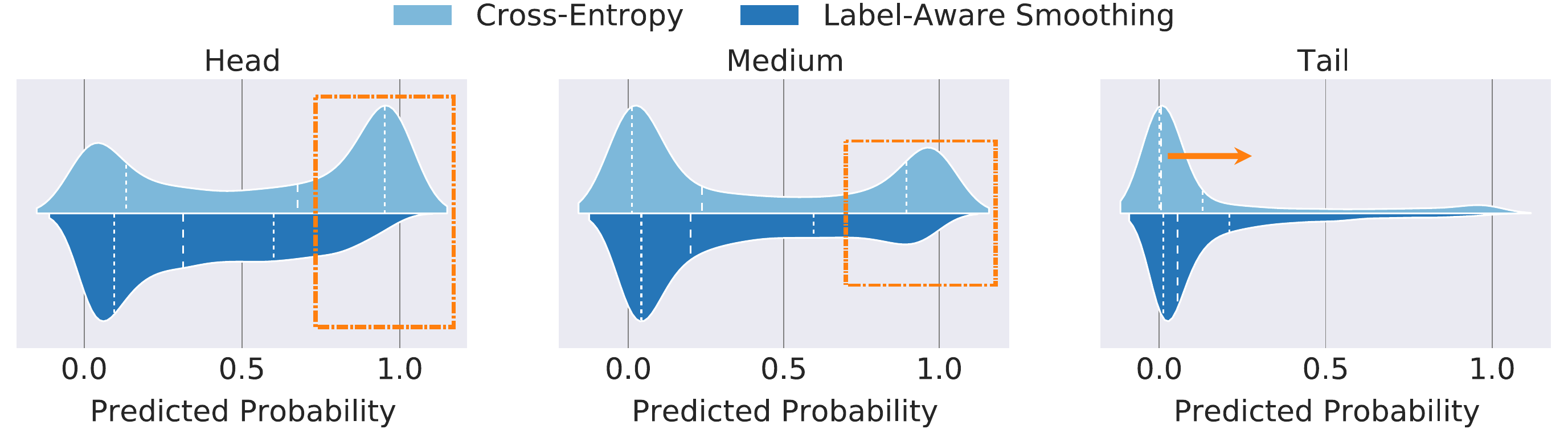} 
		\caption{Violin plot of predicted probability distributions for different parts of the classes, head (100+ images per class), medium (20-100 images per class), and tail (less than 20 images per class) on CIFAR-100-LT with IF 100. The upper half part in light blue denotes ``LWS + cross-entropy''. The bottom half part in deep blue represents ``LWS + label-aware smoothing''.}
		\label{fig:smooth}
		\vspace{-5pt}
	\end{figure*}

	\subsection{Study of mixup Strategy}\label{sec:mixup}
	
	For the two-stage learning framework, Kang \etal~\cite{decouple} and Zhou \etal~\cite{bbn} found that instance-balanced sampling gives the most general representation among all for long-tailed recognition. Besides, Thulasidasan \etal~\cite{mixup_ca} show that networks trained with mixup are better calibrated. Based on these findings, when using instance-balanced sampling, we explore the effect of mixup in the two-stage decoupling framework for higher representation generalization and over-confidence reduction.
	
	We train a plain cross-entropy model, and two two-stage models of cRT and LWS, on ImageNet-LT for 180 epochs in Stage-1 and finetune them for 10 epochs in Stage-2, respectively. We vary the training setup (with/without mixup $\alpha=0.2$) for both stages. Top-1 accuracy of these variants is listed in Table~\ref{tab:augmentation}. It reveals the following. 	
	(\romannumeral1) 
	When applying mixup, improvement of CE can be ignored. But the performance is greatly enhanced for both cRT and LWS. (\romannumeral2)~Applying additional mixup in Stage-2 yields no obvious improvement or even damages performance. The reason is that mixup encourages representation learning and is yet with adverse or negligible effect on classifier learning.  

	Besides, we draw the final classifier weight norms of these variants in Fig.~\ref{fig:norm}. We show the $L_2$ norms of the weight vectors for all classes, as well as the training data distribution sorted in a descending manner concerning the number of instances. We observe that when applying mixup (in orange), the weight norms of the tail classes tends to be large and the weight norms of the head classes decrease. It means mixup may be more friendly to tail classes.
	
	We also list ECEs of the above models in Table~\ref{tab:augmentation}. When adding mixup in just Stage-1, both cRT and LWS models can consistently obtain better top-1 accuracy and lower ECEs for different backbones (Row-4 and Row-6).  Due to the unsatisfied top-1 accuracy enhancement and unstable ECE decline of mixup for classifier learning (by adding mixup in Stage-2), we propose a label-aware smoothing to further improve both calibration and classifier learning.
	

	\subsection{Label-aware Smoothing}\label{sec:las}

	In this subsection, we analyze and deal with the two issues of over-confidence and limited improvement by classifier learning. Suppose weight of the classifier is $\displaystyle \mW \in \displaystyle \R^{M \times K}$, where $M$ is the number of features and $K$ is the number of classes. The cross-entropy encourages the whole network to be over-confident on the head classes. The cross-entropy loss after the softmax activation is $l(y, \displaystyle \vp) = -\log(\displaystyle \vp_y) = -\displaystyle \vw_y^\top\vx + \log(\sum\exp(\vw_i^\top\vx))$, where $y \in \{1, 2, ..., K\}$ is the label. $\vx \in \R^{M} $ is the feature vector send to classifier and $\vw_i$ is the $i$-th column vector of $\mW$. The optimal solution is ${\vw_y^{*}}^\top\vx = \inf$, while other $\vw_i^\top\vx$, $i\not=y$ are small enough. 
	
	Because the head classes contain much more training examples, the network makes the weight norm $\|\vw\|$ of the head classes larger to approach the optimal solution. It results in predicted probabilities mainly near 1.0 (see Fig.~\ref{fig:smooth}, the upper half in light blue). Another fact is that distributions of predicted probability are related to instance numbers. Unlike balanced recognition, applying different strategies for these classes is necessary for solving the long-tailed problem.

	
	Here, we propose label-aware smoothing to solve the over-confidence in cross-entropy and varying distributions of predicted probability issues. It is expressed as
	\begin{equation}
	\label{eqn:las}
	\begin{split}
	l(\vq, \vp) = -\sum_{i=1}^{K} \vq_i\log\vp_i, \quad \quad \quad \quad \\ 
	\vq_i=\left\{
	\begin{array}{ll}
	1-\epsilon_y = 1 - f(N_y), \quad & i= y,\\
	\frac{\epsilon_y}{K-1} = \frac{f(N_y)}{K-1}, & {\text{otherwise,}}\\
	\end{array}\right.
	\end{split}
	\end{equation}
	where $\epsilon_y$ is a small label smoothing factor for Class-$y$, relating to its class number $N_y$. Now the optimal solution becomes (proof presented in Appendix~\ref{sec:proof})
	\begin{equation}
	\label{eqn:solution_las}
	{\vw_i^*}^\top\vx=\left\{
	\begin{array}{ll}
	\log\left(\frac{(K-1)(1-\epsilon_y)}{\epsilon_y}\right) + c, \quad & i = y,\\
	c, & {\text{otherwise,}}\\
	\end{array}\right.
	\end{equation}
	where $c$ is an arbitrary real number. Compared with the optimal solution in cross-entropy, the label-aware smoothing encourages a finite output, more general and remedying overfit. We suppose the labels of the long-tailed dataset are assigned in a descending order concerning the number of instances, \ie, $N_1 \geq N_2 \geq ... \geq N_K$. Because the head classes contain more diverse examples, the predicted probabilities are more promising than those of tail classes. Thus, we require the classes with larger instance numbers to be penalized with stronger label smoothing factors -- that is, the related function $f(N_y)$ should be negatively correlated to $N_y$. We define three types of related function $f(N_y)$ as
	\begin{itemize}
		\item Concave form:
	\end{itemize}
	\begin{equation}
	\label{eqn:epsilon1}
	f(N_y) = \epsilon_K + (\epsilon_1 - \epsilon_K) \sin\left[\frac{\pi (N_y - N_K)}{2(N_1 - N_K)}\right];  \tag{3.a}
	\end{equation}
	\begin{itemize}
		\item Linear form:
	\end{itemize}
	\begin{equation}
	\label{eqn:epsilon2}
	f(N_y) = \epsilon_K + (\epsilon_1 - \epsilon_K) \frac{N_y - N_K}{N_1 - N_K}; \tag{3.b}
	\end{equation}
	\begin{itemize}
		\item Convex form:
	\end{itemize}
	\begin{equation}
	\label{eqn:epsilon3}
	f(N_y) = \epsilon_1 + (\epsilon_1 - \epsilon_K) \sin\left[\frac{3\pi}{2} + \frac{\pi (N_y - N_K)}{2(N_1 - N_K)}\right], \tag{3.c}
	\end{equation}
	where $\epsilon_1$ and $\epsilon_K$ are two hyperparameters. Illustration of these functions is shown in Fig.~\ref{fig:f_curve}. If we set $\epsilon_1 \geq \epsilon_K$, $\epsilon_1 \geq \epsilon_2 \geq ... \geq \epsilon_K$ is obtained. For large instance number $N_y$ for Class-$y$, label-aware smoothing allocates a strong smoothing factor. It lowers the fitting probability to relieve over-confidence because the head and medium classes are more likely to be over-confident than the tail classes (see Fig.~\ref{fig:smooth}).
	

	As the form of label-aware smoothing is more complicated than cross-entropy, we propose a generalized classifier learning framework to fit it. Here we give a quick review about cRT and LWS. cRT learns a classifier weight, which contains $KM$ learnable parameters, while LWS is restricted to learning the weight scaling vector $\displaystyle \vs \in \displaystyle \R^{K}$ with only $K$ learnable parameters. 
	
	In contrast, cRT has more learnable parameters and more powerful representation ability. LWS tends to obtain better validation losses and performance on large-scale datasets (refer to the experiment part in ~\cite{decouple}). So LWS has a better generalization property. To combine the advantages of cRT and LWS, we design the classifier framework in Stage-2 as
	\begin{equation}
	\label{eqn:lass}
	\vz = \text{diag}(\vs) \left(r\mW + \Delta\mW\right)^\top\vx. \tag{4}
	\end{equation}
	In Eq.~(\ref{eqn:lass}), we fix the original classifier weight $\mW$ in Stage-2. If we make the learnable scaling vector $\vs$ fixed, set $\vs = \vone$ and retention factor $r=0$, and just learn the new classifier weight $\Delta\mW \in \R^{M \times K}$, Eq.~(\ref{eqn:lass}) degrades to cRT. 
	
	Because LWS fixes the original classifier weights $\mW$ and only learns the scaling  $\vs$, Eq.~(\ref{eqn:lass}) degrades to LWS if we set $r=1$ and $\Delta\mW = \vzero$. In most cases, LWS achieves better results on large-scale datasets. Thus, we let $\vs$ learnable and set $r=1$. We also make $\Delta\mW$ learnable to improve the representation ability and optimize $\Delta\mW$ by a different learning rate. $\Delta\mW$ can be viewed as shift transformation on $\mW$. It changes the direction of weight vector $\vw$ in $\mW$, which LWS does not similarly achieve.
	
	
	\subsection{Shift Learning on Batch Normalization}
	
	In the two-stage training framework, models are first trained with instance-balanced sampling in Stage-1 and then trained with class-balanced sampling in Stage-2. Since the framework involves two samplers, or two datasets -- instance-balanced dataset $\gD_{\rm{I}}$ and class-balanced dataset $\gD_{\rm{C}}$ -- we regard this two-stage training framework as a variant of transfer learning. If we view the two-stage decoupling training framework from the transfer learning perspective, fixing the backbone part and just tuning the classifier in Stage-2 are clearly unreasonable, especially for the batch normalization~(BN) layers.
	
	Specifically, we suppose the input to network is $\vx_i$, the input feature of some BN layer is $g(\vx_i)$, and the mini-batch size is $m$. The mean and running variance of Channel-$j$ for these two stages are
	\begin{equation}\label{eqn:ib}
	\tag{5}
	\begin{split}
	\vx_i \sim P_{\gD_{\rm{I}}}(\vx, y), \quad	\vmu_{\rm{I}}^{(j)} = \frac{1}{m}\sum_{i=1}^{m}{g(\vx_i)}^{(j)},\\ 
	{{\bm{\sigma}}^{2}_{\rm{I}}}^{(j)} = \frac{1}{m} \sum_{i=1}^{m}\left[{g(\vx_i)}^{(j)} - \vmu_{\rm{I}}^{(j)}\right]^2, \quad  
	\end{split}
	\end{equation}
	\begin{equation}\label{eqn:cb}
	\tag{6}
	\begin{split}
	\vx_i \sim P_{\gD_{\rm{C}}}(\vx, y), \quad  \vmu_{\rm{C}}^{(j)} = \frac{1}{m}\sum_{i=1}^{m}{g(\vx_i)}^{(j)},\\
	\quad {{\bm{\sigma}}^{2}_{\rm{C}}}^{(j)} = \frac{1}{m} \sum_{i=1}^{m}\left[{g(\vx_i)}^{(j)} - \vmu_{\rm{C}}^{(j)}\right]^2. \quad 
	\end{split}
	\end{equation}
	
	%
	
	Due to different sampling strategies, the composition ratios of head, medium, and tail classes are also different, which lead to $P_{\gD_{\rm{I}}}(\vx, y) \not=P_{\gD_{\rm{C}}}(\vx, y)$. By Eqs.~(\ref{eqn:ib}) and (\ref{eqn:cb}), there exist biases in $\vmu$ and $\bm{\sigma}$ under two sampling strategies, \ie, $\vmu_{\rm{I}} \not= \vmu_{\rm{C}}$ and ${{\bm{\sigma}}^{2}_{\rm{I}}} \not= {{\bm{\sigma}}^{2}_{\rm{C}}}$.  Thus, it is infeasible for the decoupling framework that BN shares mean and variance across datasets with two sampling strategies. Motivated by AdaBN~\cite{li2018adaptive} and TransNorm~\cite{wang2019transferable}, we update the running mean $\vmu$ and variance $\bm{\sigma}$ and yet fix the learnable linear transformation parameters $\bm{\alpha}$ and $\bm{\beta}$ for better normalization in Stage-2.

	\begin{figure*}[t!]
		\begin{subfigure}{0.245\linewidth}
			\begin{center}
				{\quad \quad \hspace{3pt}  mixup + cRT}
			\end{center}
		\end{subfigure}
		\hfill
		\begin{subfigure}{0.245\linewidth}
			\centering
			{\quad \quad \hspace{-3pt}  mixup + LWS}
		\end{subfigure}
		\hfill
		\begin{subfigure}{0.245\linewidth}
			\centering
			{\quad {mixup + LWS + shifted BN}}
		\end{subfigure} 
		\hfill
		\begin{subfigure}{0.245\linewidth}
			\centering
			{\ MiSLAS}
		\end{subfigure} \\
		\centering
		\includegraphics[width=\textwidth]{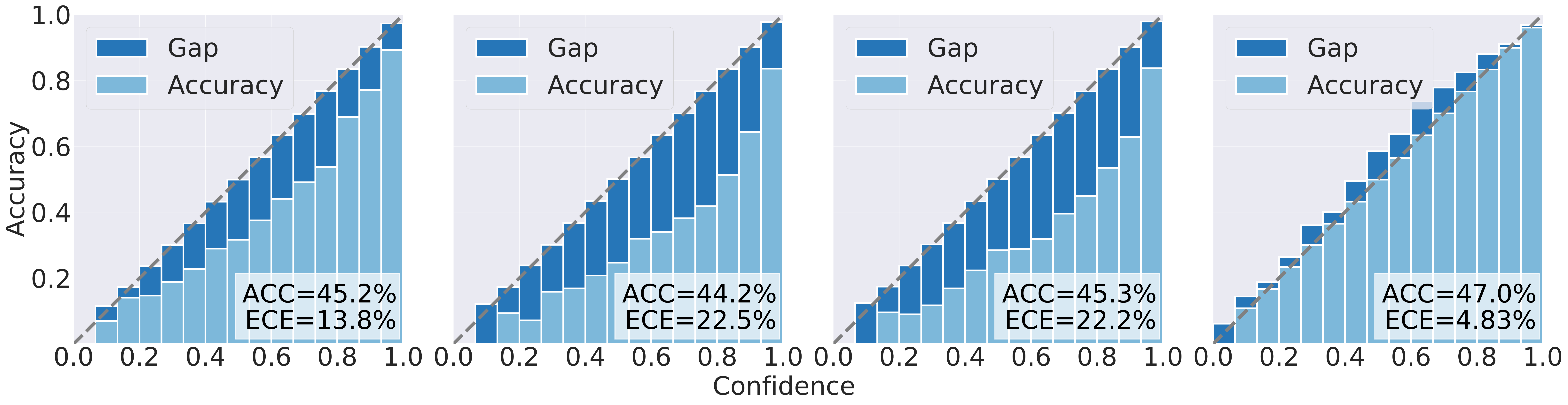} \\
		\caption{Reliability diagrams of ResNet-32 trained on CIFAR-100-LT with IF 100. From left to right: cRT with mixup, LWS with mixup, LWS with mixup and shifted BN, and MiSLAS (\textit{complying with Fig.~\ref{fig:overcondifence}}).}
		\label{fig:cifar100_rd_improve}
		\vspace{-8pt}
	\end{figure*}
	
	\section{Experiments}
	
	\vspace{1pt}
	
	\subsection{Datasets and Setup}
	
	Our experimental setup including the implementation details and evaluation protocol mainly follows \cite{ldam} for CIFAR-10-LT and CIFAR-100-LT, and \cite{decouple} for ImageNet-LT, Places-LT, and iNuturalist 2018. Please see Appendix~\ref{sec:es} for more details of training and hyperparameter setting.
	
	\vspace{-2pt}
	\subsubsection{Datasets Explanation} 
	\vspace{1pt}
	
	\noindent\textbf{CIFAR-10 and CIFAR-100.} \quad CIFAR-10 and CIFAR-100 both have 60,000 images, 50,000 for training and 10,000 for validation with 10 categories and 100 categories. For fair comparison, we use the long-tailed versions of CIFAR datasets with the same setting as those used in \cite{ldam}. It is by controlling the degrees of data imbalance with an imbalanced factor $\beta= \frac{N_{\max}}{N_{\min}}$, where $N_{\max}$ and $N_{\min}$ are the numbers of training samples for the most and the least frequent classes. Following Cao \etal~\cite{ldam} and Zhou \etal~\cite{bbn}, we conduct experiments with IF 100, 50, and 10. 
	 
	\vspace{5pt}
	
	\noindent\textbf{ImageNet-LT and Places-LT.} \quad ImageNet-LT and Places-LT were proposed by Liu \etal~\cite{liu2019large}. ImageNet-LT is a long-tailed version of the large-scale object classification dataset ImageNet~\cite{imagenet} by sampling a subset following the Pareto distribution with power value $\alpha=6$. It contains 115.8K images from 1,000 categories, with class cardinality ranging from 5 to 1,280. Places-LT is a long-tailed version of the large-scale scene classification dataset Places~\cite{places}. It consists of 184.5K images from 365 categories with class cardinality ranging from 5 to 4,980. 
	
	\vspace{5pt}
	
	\noindent\textbf{iNaturalist 2018.} \quad iNaturalist 2018~\cite{van2018inaturalist} is a classification dataset, which is on a large scale and suffers from extremely imbalanced label distribution. It is composed of 437.5K images from 8,142 categories. In addition, on iNaturalist 2018 dataset, we also face the fine-grained problem.

	\vspace{-2pt}
	\subsubsection{Implementation Details} 
	
	For all experiments, we use the SGD optimizer with momentum 0.9 to optimize networks. For CIFAR-LT, we mainly follow Cao \etal~\cite{ldam}. We train all MiSLAS models with the ResNet-32 backbone on one GPU and use the multistep learning rate schedule, which decreases the learning rate by 0.1 at the $160^{\rm{th}}$ and $180^{\rm{th}}$ epochs in Stage-1. For ImageNet-LT, Places-LT, and iNaturalist 2018, we mainly follow Kang \etal~\cite{decouple} and use the cosine learning rate schedule~\cite{loshchilov2016sgdr} to train all MiSLAS models with the ResNet-10, 50, 101, and 152 backbones on four GPUs.
	
	

	\subsection{Ablation Study}\label{sec:as}
	
	\vspace{1pt}
	
	\noindent\textbf{Calibration performance.} \quad Here we show the reliability diagrams with 15 bins of our methods on CIFAR-100-LT with IF 100 in Fig.~\ref{fig:cifar100_rd_improve}. Comparing with Fig.~\ref{fig:overcondifence}, both mixup and label-aware smoothing can not only largely enhance the network calibration (even lower ECEs than those on balanced datasets) but also greatly improve the performance for long-tailed recognition. The similar trends can also be noticed on CIFAR-10-LT, ImageNet-LT, and Places-LT (see Table~\ref{tab:augmentation} and figures in Appendix~\ref{sec:cb} for detail), which proves the power of the proposed method on calibration.
	
	All experimental results show that the training networks on imbalanced datasets lead to severe over-confidence. Since the conventional mixup and label-smoothing both contain the operation of softening the ground truth labels, it may suggest that training with hard labels is likely to be another contributing factor leading to network over-confidence.
	
	\vspace{6pt}
	
	\noindent\textbf{Comparing re-weighting with label-aware smoothing.} \quad Here we compare the proposed label-aware smoothing~(LAS) with the re-weighting methods. The main difference is on label transformation. In particular, label-aware smoothing changes the hard label to the soft version based on label distribution (see the otherwise case of Eq.~(\ref{eqn:las}): $\boldsymbol{q}_i=\frac{f(N_y)}{K-1}$, $i \neq y$). While re-weighting methods do not contain such critical transformation and just set the values to zero by $\boldsymbol{q}_i=0, i \neq y$. 
	
	Further, due to the transformation of labels, the optimal solution of ${\boldsymbol{w}_i^*}^{\top}\boldsymbol{x}$ in LAS becomes Eq.~(\ref{eqn:solution_las}). In contrast, the optimal solution of re-weighting is the same as that of cross-entropy ${{\boldsymbol{w}_i^*}^{\top}\boldsymbol{x}=\inf}$, which cannot properly change the predicted distribution and leads to over-confidence. Based on our experimental results in Table~\ref{tab:vs_reweight}, using the re-weighting method in Stage-2 degrades performance and calibration compared with the case of LAS.
	
	\begin{table}[t]
		
		\begin{center}
			\setlength{\tabcolsep}{6.5pt}
			\renewcommand\arraystretch{0.98}
			\begin{tabular}{l|ccc}
				\toprule[1.5pt]
				\textbf{Method}              & 100 & 50&  10\\ 
				\midrule
				CB-CE~\cite{effnum}  & 44.3 / 20.2 & 50.5 / 19.1 & 62.5 / 13.9 \\
				\multicolumn{1}{>{\columncolor{myblue}[6.5pt][174pt]}l|}{LAS}  & \textbf{47.0} / \textbf{4.83} & \textbf{52.3} / \textbf{2.25} & \textbf{63.2} / \textbf{1.73} \\ 
				\bottomrule[1.5pt]
			\end{tabular}
		\end{center}
		\vspace{-10pt}
		\caption{Comparison in terms of test accuracy (\%) / ECE (\%) of label-aware smoothing~(LAS) with re-weighting, class-balanced cross-entropy~(CB-CE, \cite{effnum}) in Stage-2. Both models are based on ResNet-32 and trained on CIFAR-100-LT with IF 100, 50, and 10.}
		\vspace{-16pt}
		\label{tab:vs_reweight}
	\end{table}
	
	\begin{figure*}[t]
		\begin{minipage}{0.65\textwidth}
			\begin{minipage}{0.5\textwidth}
				\centering
				\includegraphics[width=0.99\textwidth]{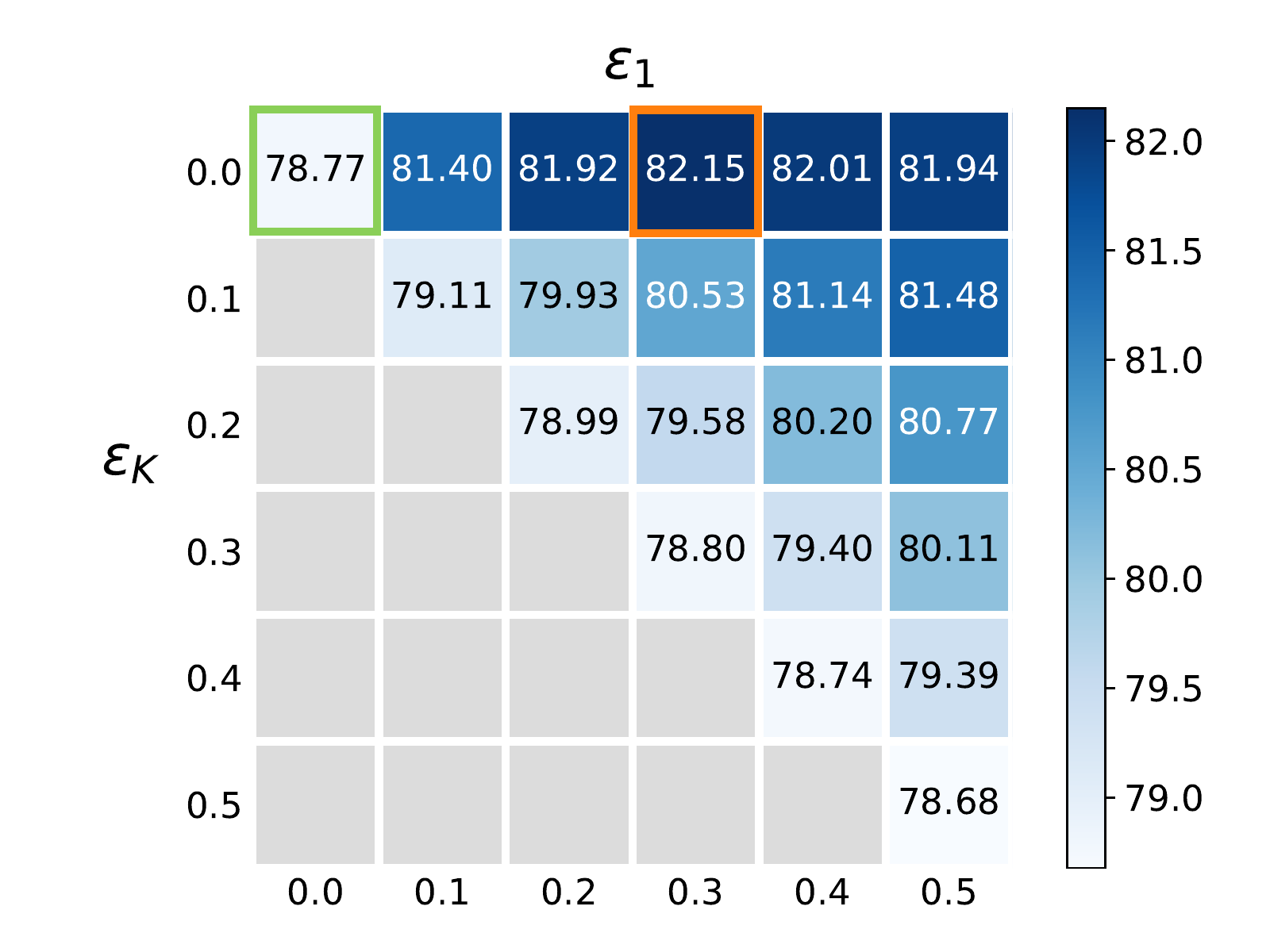} 
			\end{minipage}
			\hfill
			\begin{minipage}{0.5\textwidth}
				\centering
				\includegraphics[width=0.99\textwidth]{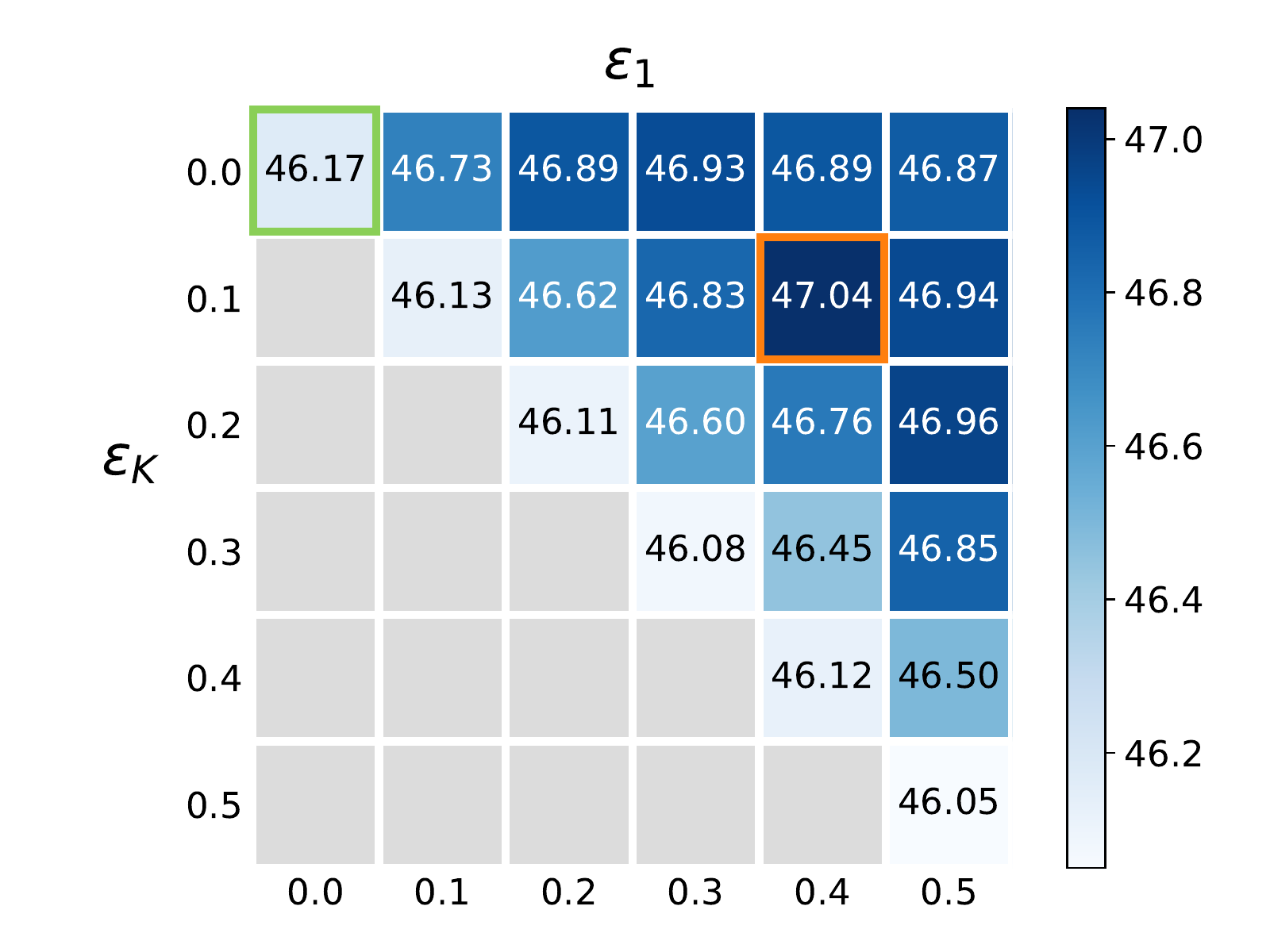} 
			\end{minipage}
			\vspace{-3.5pt}
			\caption{Ablation study of two hyperparameters $\epsilon_1$ and $\epsilon_K$ in label-aware smoothing. Heat map visualization on CIFAR-10-LT with IF 100 (left) and on CIFAR-100-LT with IF 100 (right).}
			\label{fig:as_epsilon_cifar}
		\end{minipage}
		\hfill
		\begin{minipage}{0.33\textwidth}
			\centering 
			\vspace{4pt}
			\includegraphics[width=0.98\textwidth]{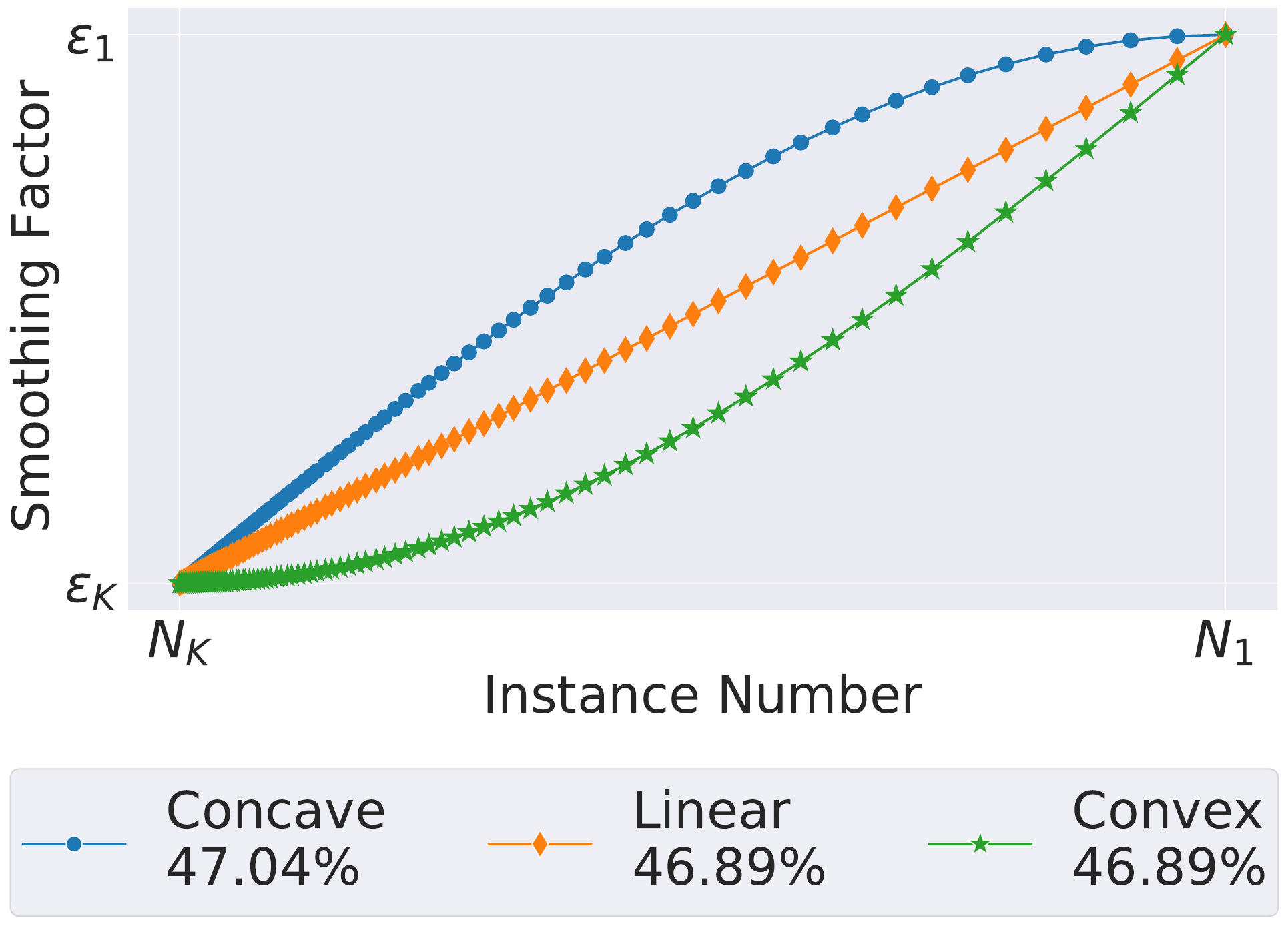}
			\vspace{-1.2pt}
			\caption{Function illustration and the test performance of Eqs.~(\ref{eqn:epsilon1}), (\ref{eqn:epsilon2}), and (\ref{eqn:epsilon3}). Concave form achieves the best result.} 
			\label{fig:f_curve}
			\vspace{-4pt}
		\end{minipage}
	\end{figure*}
	
	\begin{figure*}[t]	
		\begin{minipage}{0.65\textwidth}
			\centering 
			\vspace{1pt}
			\includegraphics[width=0.99\textwidth]{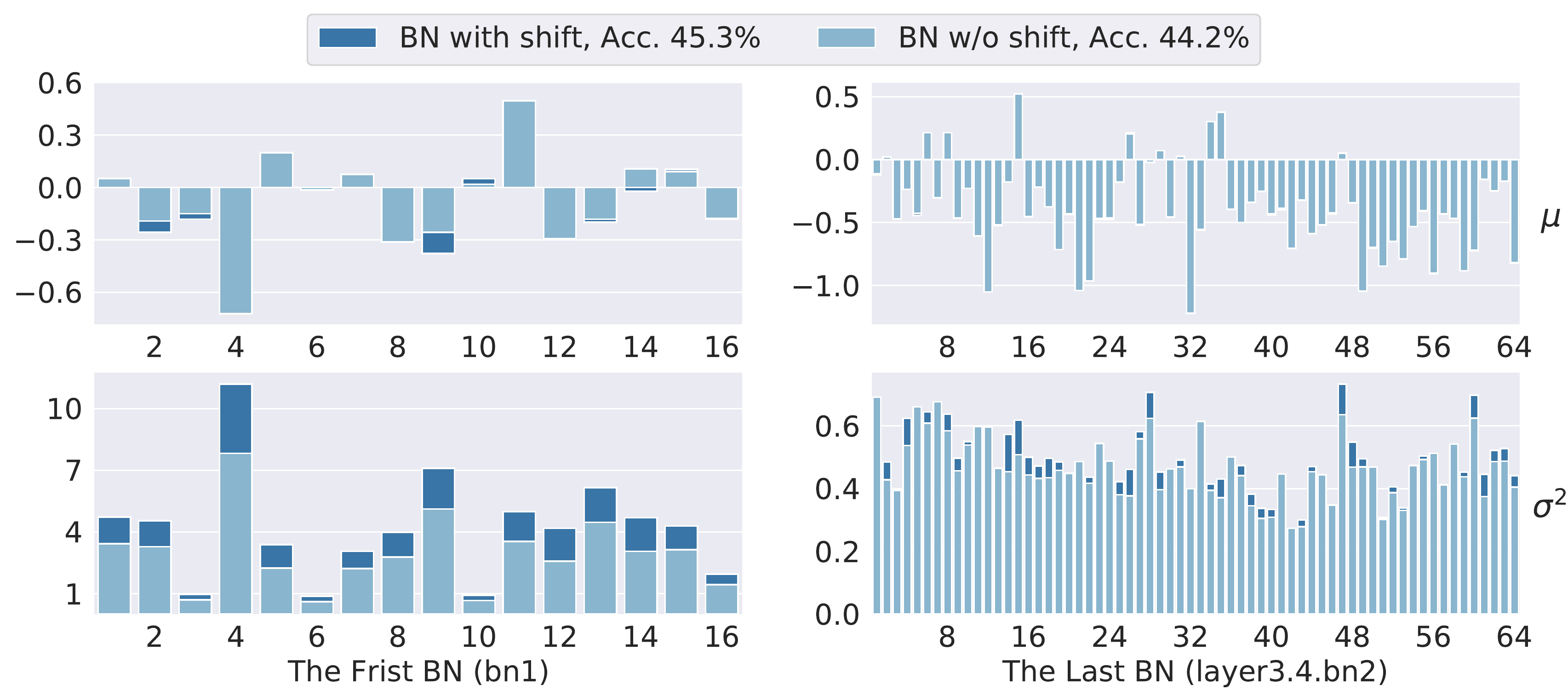}
			\vspace{2pt}
			\caption{Visualization of the changes in the running mean $\vmu$ and variance ${\bm{\sigma}}^2$. The ResNet-32 based model is trained on CIFAR-100-LT with IF 100. Left: $\vmu$ and ${\bm{\sigma}}^2$ in the first BN of ResNet-32, which contains 16 channels. Right: $\vmu$ and ${\bm{\sigma}}^2$ in the last BN of ResNet-32, which contains 64 channels.} 
			\label{fig:bn_fix_move}
		\end{minipage}
		\hspace{10pt}
		\begin{minipage}{0.33\textwidth}
			\setlength{\tabcolsep}{5.1pt}
			\begin{center}
				\vspace{-4pt}
				\renewcommand\arraystretch{0.9}
				\begin{tabular}{ccc|ccc}
					\toprule[1.5pt]
					\multicolumn{3}{c|}{Module}	  & \multicolumn{3}{c}{CIFAR-100-LT} \\ 
					\cmidrule(lr){1-3}\cmidrule(lr){4-6}
					MU & SL &  LAS   & 100 & 50 & 10 \\ 
					\midrule
					\rlap{\raisebox{0.3ex}{\hspace{0.4ex}\scriptsize \ding{56}}}$\square$ & \rlap{\raisebox{0.3ex}{\hspace{0.4ex}\scriptsize \ding{56}}}$\square$ & \rlap{\raisebox{0.3ex}{\hspace{0.4ex}\scriptsize \ding{56}}}$\square$  
					& 41.2  & 46.0  &  58.5 \\
					\rlap{\raisebox{0.3ex}{\hspace{0.4ex}\tiny \ding{52}}}$\square$ & \rlap{\raisebox{0.3ex}{\hspace{0.4ex}\scriptsize \ding{56}}}$\square$ & \rlap{\raisebox{0.3ex}{\hspace{0.4ex}\scriptsize \ding{56}}}$\square$ & 44.2  & 50.6  &  62.2  \\
					\rlap{\raisebox{0.3ex}{\hspace{0.4ex}\tiny \ding{52}}}$\square$ & \rlap{\raisebox{0.3ex}{\hspace{0.4ex}\tiny \ding{52}}}$\square$ & \rlap{\raisebox{0.3ex}{\hspace{0.4ex}\scriptsize \ding{56}}}$\square$ & 45.3  & 51.4  &  62.8   \\
					\multicolumn{1}{>{\columncolor{myblue}[5pt][139.5pt]}c}{\rlap{\raisebox{0.3ex}{\hspace{0.4ex}\tiny \ding{52}}}$\square$} & \rlap{\raisebox{0.3ex}{\hspace{0.4ex}\tiny \ding{52}}}$\square$ & \rlap{\raisebox{0.3ex}{\hspace{0.4ex}\tiny \ding{52}}}$\square$ & \textbf{47.0} & \textbf{52.3}  &  \textbf{63.2}   \\
					\midrule
					\rlap{\raisebox{0.3ex}{\hspace{0.4ex}\scriptsize \ding{56}}}$\square$ & \rlap{\raisebox{0.3ex}{\hspace{0.4ex}\scriptsize \ding{56}}}$\square$ & \rlap{\raisebox{0.3ex}{\hspace{0.4ex}\scriptsize \ding{56}}}$\square$  
					& 36.3  & 34.2  &  27.5\\
					\rlap{\raisebox{0.3ex}{\hspace{0.4ex}\tiny \ding{52}}}$\square$ & \rlap{\raisebox{0.3ex}{\hspace{0.4ex}\scriptsize \ding{56}}}$\square$ & \rlap{\raisebox{0.3ex}{\hspace{0.4ex}\scriptsize \ding{56}}}$\square$ & 22.5  & 18.4  &  14.3  \\
					\rlap{\raisebox{0.3ex}{\hspace{0.4ex}\tiny \ding{52}}}$\square$ & \rlap{\raisebox{0.3ex}{\hspace{0.4ex}\tiny \ding{52}}}$\square$ & \rlap{\raisebox{0.3ex}{\hspace{0.4ex}\scriptsize \ding{56}}}$\square$ & 22.2  & 19.2  &  13.7   \\
					
					\multicolumn{1}{>{\columncolor{myblue}[5pt][139.5pt]}c}{\rlap{\raisebox{0.3ex}{\hspace{0.4ex}\tiny \ding{52}}}$\square$} & \rlap{\raisebox{0.3ex}{\hspace{0.4ex}\tiny \ding{52}}}$\square$ & \rlap{\raisebox{0.3ex}{\hspace{0.4ex}\tiny \ding{52}}}$\square$ & \textbf{4.83} & \textbf{2.25}  &  \textbf{1.73}   \\
					\bottomrule[1.5pt]     
				\end{tabular}
				\vspace{-4pt}
				\captionof{table}{Ablation study for all proposed modules on CIFAR-100-LT. Top: accuracy (\%). Bottom: ECE (\%). MU: applying mixup only in Stage-1. SL: shift learning on BN. LAS: label-aware smoothing.}
				\vspace{-8pt}
				\label{table:as}
			\end{center}
		\end{minipage}
		\vspace{-10pt}
	\end{figure*}

	\vspace{6pt}

	\noindent\textbf{How $\epsilon_1$ and $\epsilon_K$ affect label-aware smoothing?} \quad In our label-aware smoothing, there are two hyperparameters in Eqs. (\ref{eqn:epsilon1}), (\ref{eqn:epsilon2}), and (\ref{eqn:epsilon3}). \!They are $\epsilon_1$ and $\epsilon_K$, which control penalty of classes. In a recognition system, if the predicted probability of Class-$y$ is larger than 0.5, the classifier would classify the input to Class-$y$. Thus, to make it reasonable, we limit $0 \leq\epsilon_K \leq\epsilon_1 \leq 0.5$. 
	
	Here we conduct experiments by varying $\epsilon_1$ and $\epsilon_K$ both from 0.0 to 0.5 on CIFAR-10-LT with IF 100. We plot the performance matrix upon $\epsilon_1$ and $\epsilon_K$ in Fig.~\ref{fig:as_epsilon_cifar} for all possible variants. It shows that the classification accuracy is further improved by 3.3\% comparing with conventional cross-entropy ($\epsilon_1=0$ and $\epsilon_K=0$, green square) when we pick $\epsilon_1=0.3$, and $\epsilon_K=0.0$ (orange square) for label-aware smoothing. Consistent improvement 0.9\% is yielded on CIFAR-100-LT with IF 100 when picking $\epsilon_1=0.4$ and $\epsilon_K=0.1$ for label-aware smoothing. 
	
	\vspace{6pt}
	
	\noindent\textbf{How $f(\cdot)$ affects label-aware smoothing?} \quad As discussed in Sec.~\ref{sec:las}, the related function $f(\cdot)$ may play a significant role for the final model performance. We draw illustration of Eqs.~(\ref{eqn:epsilon1}), (\ref{eqn:epsilon2}), and (\ref{eqn:epsilon3}) in Fig.~\ref{fig:f_curve}. For CIFAR-100-LT with IF 100, we set $K=100$, $N_1=500$, and $N_{100}=5$. Based on the ablation study results of $\epsilon_1$ and $\epsilon_K$ mentioned above, we set $\epsilon_1=0.4$ and $\epsilon_{100}=0.1$ here. After tuning for 10 epochs in Stage-2, accuracy of the concave model is the best. We also exploit other forms, \eg, exponential form of $f(\cdot)$, in Appendix~\ref{sec:as_form}. The gain of changing form is quite limited compared with varying $\epsilon_1$ and $\epsilon_K$.

	\vspace{6pt}
	
	\noindent\textbf{How label-aware smoothing affects prediction distribution?} \quad To visualize the change in predicted probability distributions, we train two LWS models, one with cross-entropy and the other with label-aware smoothing on CIFAR-100-LT with IF 100. The cross-entropy-based distributions of the head, medium, and tail classes are shown in the upper part of Fig.~\ref{fig:smooth} in light blue. The label-aware smoothing-based distributions are in the bottom half in deep blue. We observe that the over-confidence of head and medium classes is much reduced, and the whole distribution of the tail classes slightly moves right when using label-aware smoothing. These empirical results are consistent with our analysis in Sec.~\ref{sec:las}.

	\begin{table*}[t]
		\setlength{\tabcolsep}{8.8pt}
		\centering
		\begin{tabular}{l|lll|lll}
			\toprule[1.5pt]
			{\multirow{2.6}*{\textbf{Method}}}	 &\multicolumn{3}{c|}{CIFAR-10-LT} & \multicolumn{3}{c}{CIFAR-100-LT} \\ 
			\cmidrule(lr){2-4}\cmidrule(l){5-7}
			& 100 & 50 &  10 & 100 & 50 & 10 \\ 
			\midrule
			
			CE & 70.4  & 74.8  & 86.4  & 38.4  & 43.9  &  55.8   \\
			mixup~\cite{mixup} & 73.1  & 77.8 &  87.1  & 39.6  & 45.0  &  58.2   \\
			LDAM+DRW~\cite{ldam} & 77.1  & 81.1  &  88.4  & 42.1 & 46.7  &  58.8    \\
			
			BBN{\scriptsize{(include mixup)}}~\cite{bbn} & 79.9  & 82.2  &  88.4  & 42.6  & 47.1  &  59.2  \\
			

			Remix+DRW{\scriptsize{(300 epochs)}}~\cite{remix} & 79.8  & -  &  89.1  & 46.8  & -  &  61.3  \\
			\midrule
			cRT+mixup & 79.1 / 10.6  & 84.2 /  6.89  &  89.8 / 3.92 & 45.1 / 13.8 & 50.9 / 10.8 &  62.1 / 6.83  \\
			LWS+mixup & 76.3 / 15.6  & 82.6 / 11.0  &  89.6 / 5.41  & 44.2 / 22.5  & 50.7 / 19.2  &  62.3 / 13.4  \\
			
			\multicolumn{1}{>{\columncolor{myblue}[8.8pt][371.5pt]}l|}{MiSLAS}  & \textbf{82.1} / \textbf{3.70}  & \textbf{85.7} / \textbf{2.17}  &  \textbf{90.0} / \textbf{1.20}  & \textbf{47.0} / \textbf{4.83}  & \textbf{52.3} / \textbf{2.25}  &  \textbf{63.2} / \textbf{1.73}  \\
			\bottomrule[1.5pt]     
		\end{tabular}
		\caption{Top-1 accuracy (\%) / ECE (\%) for ResNet-32 based models trained on CIFAR-10-LT and CIFAR-100-LT.}
		\label{tab:cifar}
	\end{table*}
	
	\begin{table*}[t]

		\begin{minipage}[t]{0.32\textwidth}
			\setlength{\tabcolsep}{5pt}
			\begin{center}
				\begin{tabular}{l|l}
					\toprule[1.5pt]
					\textbf{Method}              & ResNet-50     \\ \midrule
					CE & 44.6         \\
					CE+DRW~\cite{ldam} & 48.5         \\
					Focal+DRW~\cite{lin2017focal}  & 47.9 \\
					LDAM+DRW~\cite{ldam}    & 48.8 \\ 
					\midrule
					CRT+mixup & 51.7 / 5.62 \\
					LWS+mixup & 52.0 / 2.23 \\ 
					\multicolumn{1}{>{\columncolor{myblue}[5pt][58pt]}l|}{MiSLAS}      & \textbf{52.7} / \textbf{1.83} \\ \bottomrule[1.5pt]
				\end{tabular}
			\end{center}
			\begin{center}

				(a) ImageNet-LT
			\end{center}
		\end{minipage}
		\hfill
		\begin{minipage}[t]{0.32\textwidth}
			\setlength{\tabcolsep}{5pt}
			\begin{center}
				\begin{tabular}{l|l}
					\toprule[1.5pt]
					\textbf{Method}              & ResNet-50 \\ \midrule
					CB-Focal~\cite{effnum}  & 61.1\\
					LDAM+DRW~\cite{ldam}  &68.0 \\ 
					BBN{\scriptsize(include mixup)}~\cite{bbn} & 69.6 \\
					Remix+DRW~\cite{remix} & 70.5  \\ 
					\midrule
					cRT+mixup     &  70.2 / \textbf{1.79} \\
					LWS+mixup{\scriptsize(under-conf.)} & 70.9 / 9.41 \\ 
					\multicolumn{1}{>{\columncolor{myblue}[5pt][58pt]}l|}{MiSLAS{\scriptsize(under-conf.)}}  & \textbf{71.6} / 7.67 \\ \bottomrule[1.5pt]
				\end{tabular}
			\end{center}
			\begin{center}
		
				(b) iNaturalist 2018
			\end{center}
		\end{minipage}
		\hfill
		\hspace{3pt}
		\begin{minipage}[t]{0.32\textwidth}
			\setlength{\tabcolsep}{5pt}
			\begin{center}
				\begin{tabular}{l|l}
					\toprule[1.5pt]
					\textbf{Method}              & ResNet-152     \\ \midrule  
					Range Loss~\cite{zhang2017range}      & 35.1 \\
					FSLwF~\cite{gidaris2018dynamic} & 34.9 \\ 
					OLTR~\cite{liu2019large} & 35.9 \\
					OLTR+LFME~\cite{xiang2020learning} & 36.2 \\
					\midrule
					cRT+mixup & 38.3 / 12.4 \\ 
					LWS+mixup & 39.7 / 11.7        \\
					\multicolumn{1}{>{\columncolor{myblue}[5pt][63pt]}l|}{MiSLAS}        & \textbf{40.4} / \textbf{3.59} \\ \bottomrule[1.5pt]
				\end{tabular}
			\end{center}
			\begin{center}

				(c) Places-LT
			\end{center}
		\end{minipage}

		\caption{Top-1 accuracy (\%) / ECE (\%) on ImageNet-LT (left), iNaturalist 2018 (center) and Places-LT (right). }
		\vspace{-5pt}
		\label{tab:large_dataset}
	\end{table*}
	
	\vspace{5pt}
	
	\noindent\textbf{Further analysis of shift learning.} \quad In this part, we conduct experiments to show the effectiveness and suitability of shift learning on BN. We train the LWS model on CIFAR-100-LT with IF 100. After 10-epoch finetuning in Stage-2, the model trained with BN shifting achieves accuracy $45.3\%$, $1.1\%$ higher than that without BN shifting. We also visualize the change in BN. As shown in Fig.~\ref{fig:bn_fix_move}, there exist biases in $\vmu$ and ${\bm{\sigma}}^2$ between datasets using different sampling strategies. 
	
	\vspace{1pt}
	
	Due to different composition ratios of the head, medium and tail classes, the statistic mean $\vmu$ and variance ${\bm{\sigma}}^2$ vary. We also notice intriguing phenomena in Fig.~\ref{fig:bn_fix_move}: (\romannumeral1) the change in variance ${\bm{\sigma}}^2$ is larger than that on mean $\vmu$. (\romannumeral2) Change of $\vmu$ and ${\bm{\sigma}}^2$ in the deep BN layers is much smaller than that in the shallow BN layers.
	
	\vspace{5pt}
	
	\noindent\textbf{Summary.} \quad  Overall, Table~\ref{table:as} shows the ablation investigation on the effects of mixup (adding mixup in Stage-1, MU), shift learning on batch normalization (SL), and label-aware smoothing (LAS). We note each proposed module can not only improves accuracy (top of Table~\ref{table:as}), but also greatly relieves over-confidence (bottom of Table~\ref{table:as}) on CIFAR-100-LT for all commonly-used imbalanced factors, \ie, 100, 50, and 10. They firmly manifest the effectiveness.

	\subsection{Comparison with State-of-the-arts}

	To verify the effectivity, we compare the proposed method against previous one-stage methods of Range Loss~\cite{zhang2017range}, LDAM Loss~\cite{ldam}, FSLwF~\cite{gidaris2018dynamic}, and OLTR~\cite{liu2019large}, and against previous two-stage methods, including DRS-like, DRW-like~\cite{ldam}, LFME~\cite{xiang2020learning}, cRT, and LWS~\cite{decouple}. For fair comparison, we add mixup on the LWS and cRT models. Remix~\cite{remix} is a recently proposed augmentation method for long-tail recognition. Because BBN~\cite{bbn} has double samplers and is trained in a mixup-like manner, we directly compare our method with it.

	\vspace{6pt}
	\noindent\textbf{Experimental results on CIFAR-LT.} \quad  We conduct extensive experiments on CIFAR-10-LT and CIFAR-100-LT with IF 100, 50, and 10, using the same setting as previous work~\cite{ldam,bbn}. The results are summarized in Table~\ref{tab:cifar}. Compared with previous methods, our MiSLAS outperforms all previous methods by consistently large margins both in top-1 accuracy and ECE. Moreover, the superiority holds for all imbalanced factors, \ie, 100, 50, and 10, on both CIFAR-10-LT and CIFAR-100-LT.

	\vspace{6pt}
	\noindent\textbf{Experimental results on large-scale datasets.} 
	\quad We further verify the effectiveness of our method on three large-scale imbalanced datasets, \ie, ImageNet-LT, iNaturalist 2018, and Places-LT. Table~\ref{tab:large_dataset} lists experimental results on ImageNet-LT (left), iNaturalist 2018 (center), and Places-LT (right). Notably, our MiSLAS outperforms other approaches and sets a new state-of-the-art with better accuracy and confidence calibration on almost all three large-scale long-tailed benchmark datasets. More results about the split class accuracy and different backbones on these three datasets are listed in Appendix~\ref{sec:more_result}.

	\section{Conclusion}
	
	\vspace{1pt}
	
	In this paper, we have discovered that models trained on long-tailed datasets are more miscalibrated and over-confident than those trained on balanced datasets. We accordingly propose two solutions of using mixup and designing label-aware smoothing to handle different degrees of over-confidence for classes. We note the dataset bias (or domain shift) in two-stage resampling methods for long-tailed recognition. To reduce dataset bias in the decoupling framework, we propose shift learning on the batch normalization layer, which further improves the performance. Extensive quantitative and qualitative experiments on various benchmarks show that our MiSLAS achieves decent performance for both top-1 recognition accuracy and confidence calibration, and makes a new state-of-the-art.
	
	\clearpage
	{   
		\small
		\bibliographystyle{ieee_fullname}
		\bibliography{cvpr}
	}
	\clearpage

	
	\appendix
	\onecolumn
	\begin{center}
		\Large \textbf{Improving Calibration for Long-Tailed Recognition (Supplementary Material)}
	\end{center}
	\section{Experiment Setup}\label{sec:es}

	Following Liu \etal~\cite{liu2019large} and Kang \etal~\cite{decouple}, we report the commonly used top-1 accuracy over all classes on the balanced test/validation datasets, denoted as \textit{All}. We further report accuracy on three splits of classes: \textit{Head-Many} (more than 100 images), \textit{Medium} (20 to 100 images), and \textit{Tail-Few} (less than 20 images). The detailed setting of hyperparameters and training for all datasets used in our paper are listed in Table~\ref{tab:setting}.
	
	\begin{table*}[h]
		\setlength{\tabcolsep}{6pt}
		\centering
		\vspace{-5pt}
		\begin{tabular}{ll|ccc|cc|ccccc}
			\toprule[1.5pt]
			{\multirow{2.5}*{\textbf{Dataset}}}	& &\multicolumn{3}{c|}{Common setting} & \multicolumn{2}{c|}{Stage-1}  & \multicolumn{5}{c}{Stage-2}\\ 
			\cmidrule(lr){3-5}\cmidrule(lr){6-7}\cmidrule(lr){8-12}
			& & LR & BS 	 & WD & {Epochs}	& LRS & {Epochs}	& LRS & $\epsilon_1$ & $\epsilon_K$ &	$\Delta\mW$ \\ 
			\midrule
			CIFAR-10-LT& $\beta=10$ & 0.1  & 128  & $2\times10^{-4}$  & 200  & multistep  &  10 & cosine & 0.1 & 0.0 &   0.2$\times$ \\
			CIFAR-10-LT& $\beta=50$  & 0.1  & 128  & $2\times10^{-4}$  & 200  & multistep  &  10 & cosine & 0.2 & 0.0 &   0.2$\times$ \\
			CIFAR-10-LT& $\beta=100$  & 0.1  & 128  & $2\times10^{-4}$  & 200  & multistep  &  10 & cosine & 0.3 & 0.0 &   0.5$\times$ \\
			\midrule
			CIFAR-100-LT& $\beta=10$  & 0.1  & 128  & $2\times10^{-4}$  & 200  & multistep  &  10 & cosine & 0.2 & 0.0 &   0.1$\times$ \\
			CIFAR-100-LT& $\beta=50$  & 0.1  & 128  & $2\times10^{-4}$  & 200  & multistep  &  10 & cosine & 0.3 & 0.0 &   0.1$\times$ \\
			CIFAR-100-LT& $\beta=100$  & 0.1  & 128  & $2\times10^{-4}$  & 200  & multistep  &  10 & cosine & 0.4 & 0.1 &   0.2$\times$ \\
			\midrule
			ImageNet-LT & & 0.1  & 256  & $5\times10^{-4}$  & 180  & cosine  &  10 & cosine & 0.3 & 0.0 &   0.05$\times$ \\
			Places-LT & & 0.1  & 256  & $5\times10^{-4}$ & 90  & {cosine}  &  10 & {cosine} & 0.4 & 0.1 &   0.05$\times$ \\
			iNaturalist 2018 &  & 0.1  & 256  & $1\times10^{-4}$  & 200  & {cosine}  &  30 & {cosine} & 0.4 & 0.0 &   0.05$\times$ \\
			\bottomrule[1.5pt]     
		\end{tabular}
		\vspace{-5pt}
		\caption{Detailed experiment setting on five benchmark datasets. LR: initial learning rate, BS: batch size, WD: weight decay, LRS: learning rate schedule, and $\Delta\mW$: learning rate ratio of $\Delta\mW$.}
		\vspace{-10pt}
		\label{tab:setting}
	\end{table*}
	
	\section{Exponential Form of the Related Function $f(\cdot)$}\label{sec:as_form}
\begin{figure}[h]
		\begin{subfigure}{0.495\linewidth}
			\centering
			\includegraphics[width=0.98\textwidth]{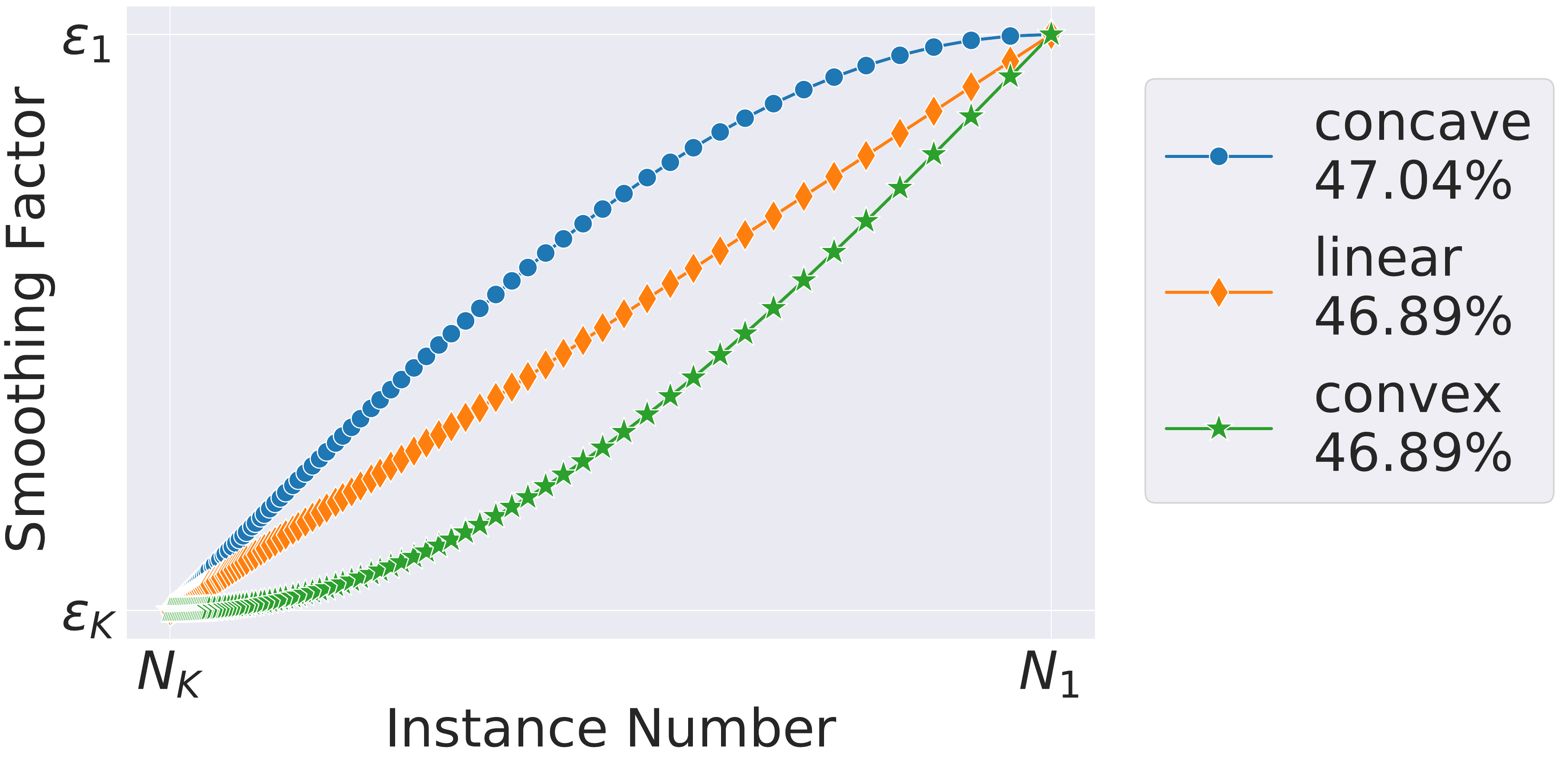}
		\end{subfigure}
		\hfill
		\begin{subfigure}{0.495\linewidth}
			\centering
			\includegraphics[width=0.98\textwidth]{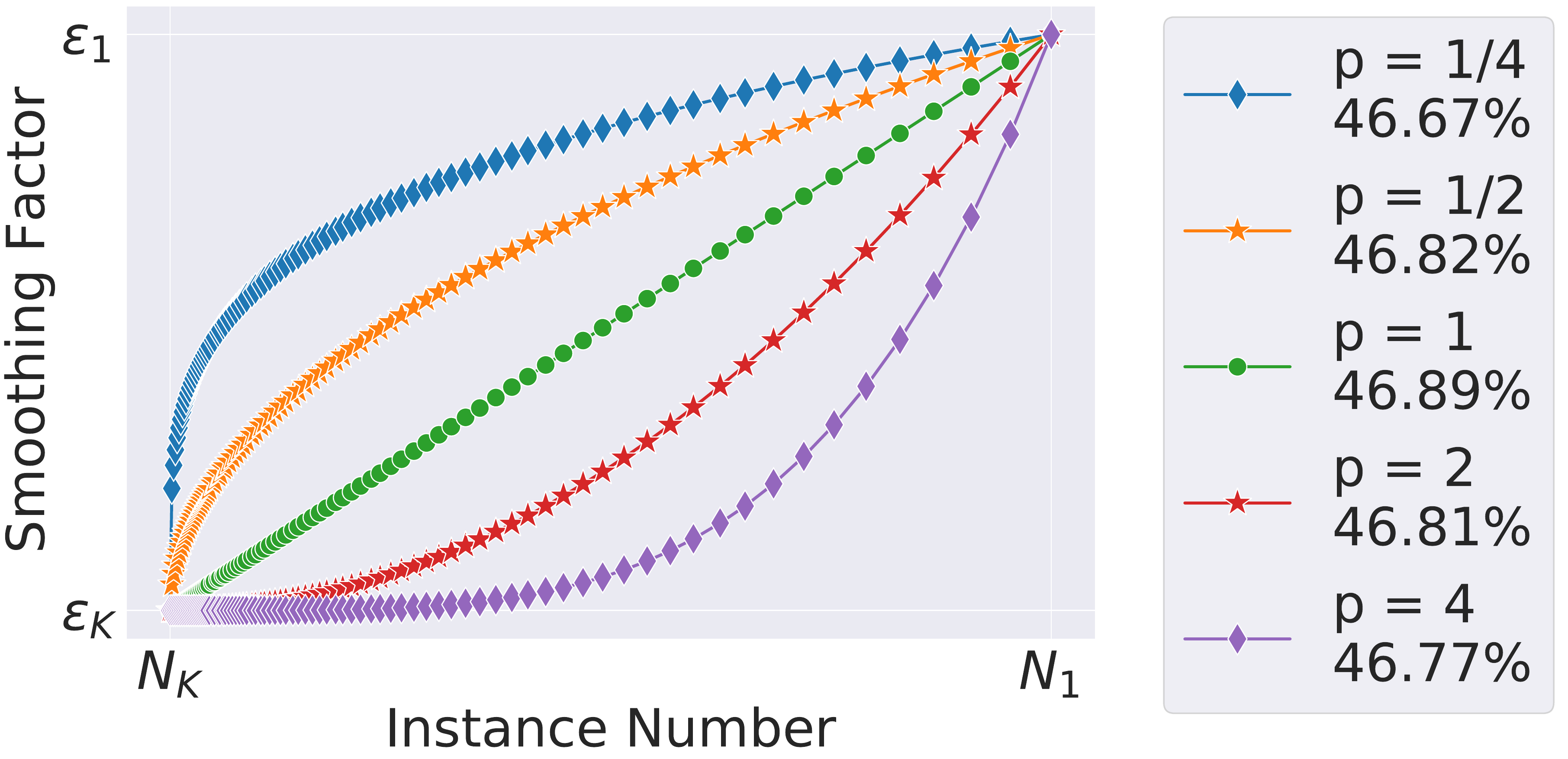}
		\end{subfigure} \\
		
		\caption{Function illustration and accuracy of Eqs.~(\ref{eqn:epsilon1}), (\ref{eqn:epsilon2}), and (\ref{eqn:epsilon3}) (left) and Eq.~(\ref{eqn:epsilon_2}) (right).}
		\label{fig:ab_curve}
	\end{figure}

	As discussed in Secs.~\ref{sec:las} and \ref{sec:as}, the form of the related function $f(\cdot)$ may play an important role for final model performance. We draw the illustration of Eqs.~(\ref{eqn:epsilon1}), (\ref{eqn:epsilon2}), and (\ref{eqn:epsilon3}) at the left of Fig.~\ref{fig:ab_curve}. For the CIFAR-100-LT dataset with imbalanced factor 100, $K=100$, $N_1=500$, and $N_{100}=5$. Based on the ablation study results of $\epsilon_1$ and $\epsilon_K$ mentioned in Sec.~\ref{sec:as}, we set $\epsilon_1=0.4$ and $\epsilon_{100}=0.1$ here. After fintuning for 10 epochs in Stage-2, the accuracy of the concave model is the best. We also design an exponential related function, which is written as
	\begin{equation}
	\label{eqn:epsilon_2}
	\epsilon_y = f(N_y) = \epsilon_K + (\epsilon_1 - \epsilon_K) {\left(\frac{N_y - N_K}{N_1 - N_K}\right)}^p, \quad\quad y = 1, 2, ..., K,\tag{7}
	\end{equation}
	where $p$ is a hyperparameter to control the shape of the related function. For example, we get the concave related function when setting $p<1$ and convex function otherwise. Illustration of Eq.~(\ref{eqn:epsilon_2}) is given on the right of Fig.~\ref{fig:ab_curve}. Comparing accuracy of all variants, the influence of the related function form is quite limited for the final performance (0.3\% increase). Because the concave related function Eq.~(\ref{eqn:epsilon1}) achieves the best performance, we choose it as the default setting of the related function $f(\cdot)$ for other experiments.
	
	\clearpage
	\section{Calibration Performance}\label{sec:cb}
	
	\begin{figure}[h]
		\begin{subfigure}{0.195\linewidth}
			\begin{center}
				{\quad \quad Org. CIFAR-10}
			\end{center}
		\end{subfigure}
		\hfill
		\begin{subfigure}{0.195\linewidth}
			\centering
			{\quad\hspace{4.5pt}  CIFAR10-LT, IF100}
		\end{subfigure}
		\hfill
		\begin{subfigure}{0.195\linewidth}
			\centering
			{\quad {cRT}}
		\end{subfigure} 
		\hfill
		\begin{subfigure}{0.195\linewidth}
			\centering
			{\hspace{2pt}  LWS}
		\end{subfigure} 
		\hfill
		\begin{subfigure}{0.195\linewidth}
			\centering
			{ MiSLAS}
		\end{subfigure}\\
		\centering
		\includegraphics[width=\textwidth]{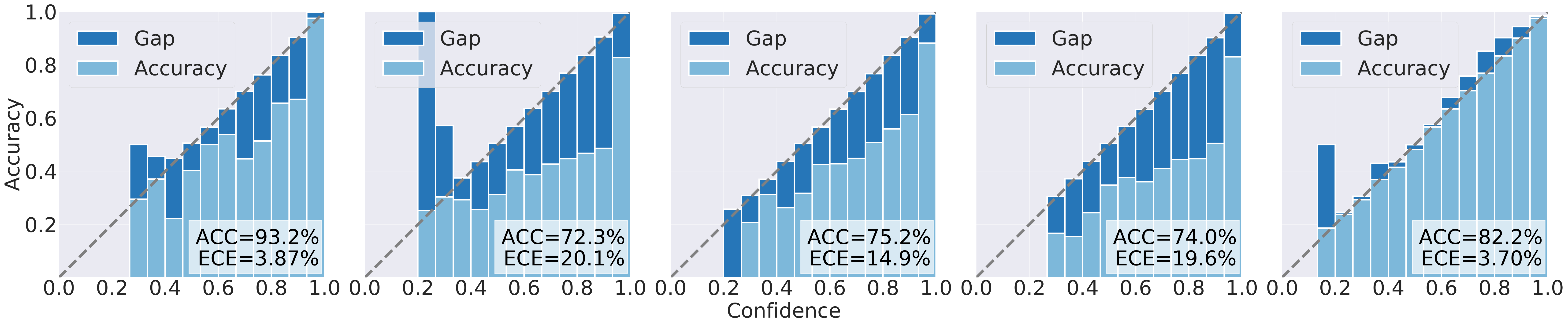} \\
		\vspace{-5pt}
		\caption{Reliability diagrams on CIFAR10 with 15 bins. From left to right: plain ResNet-32 model trained on the original CIFAR-10 dataset, plain model, cRT, LWS, and MiSLAS trained on long-tailed CIFAR-10 with imbalanced factor 100.}
		\vspace{-10pt}
		\label{fig:more_rd_cifar10}
	\end{figure}

	\begin{figure}[h]
		\begin{subfigure}{0.195\linewidth}
			\begin{center}
				{\quad \quad Org. ImageNet}
			\end{center}
		\end{subfigure}
		\hfill
		\begin{subfigure}{0.195\linewidth}
			\centering
			{\quad  \hspace{3pt}  ImageNet-LT}
		\end{subfigure}
		\hfill
		\begin{subfigure}{0.195\linewidth}
			\centering
			{\quad {cRT}}
		\end{subfigure} 
		\hfill
		\begin{subfigure}{0.195\linewidth}
			\centering
			{\hspace{2pt}  LWS}
		\end{subfigure} 
		\hfill
		\begin{subfigure}{0.195\linewidth}
			\centering
			{ MiSLAS}
		\end{subfigure}\\
		\centering
		\includegraphics[width=\textwidth]{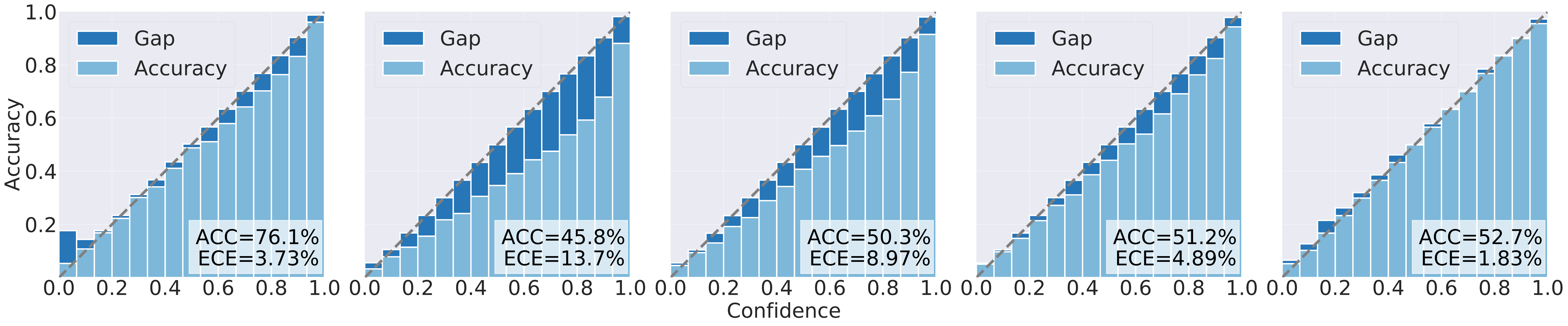}
		\vspace{-15pt}
		\caption{Reliability diagrams on ImageNet with 15 bins. From left to right: plain ResNet-50 model trained on the original ImageNet dataset, plain model, cRT, LWS, and MiSLAS trained on ImageNet-LT.}
		\vspace{-10pt}
		\label{fig:more_rd_imgnet}
	\end{figure}

	\begin{figure}[h]
		\begin{subfigure}{0.195\linewidth}
			\begin{center}
				{\quad \quad cRT}
			\end{center}
		\end{subfigure}
		\hfill
		\begin{subfigure}{0.195\linewidth}
			\centering
			{\quad   \hspace{1.5pt}  LWS}
		\end{subfigure}
		\hfill
		\begin{subfigure}{0.195\linewidth}
			\centering
			{\quad {mixup + cRT}}
		\end{subfigure} 
		\hfill
		\begin{subfigure}{0.195\linewidth}
			\centering
			{\quad \hspace{-5pt}  mixup + LWS}
		\end{subfigure} 
		\hfill
		\begin{subfigure}{0.195\linewidth}
			\centering
			{MiSLAS}
		\end{subfigure}\\
		\centering
		\includegraphics[width=\textwidth]{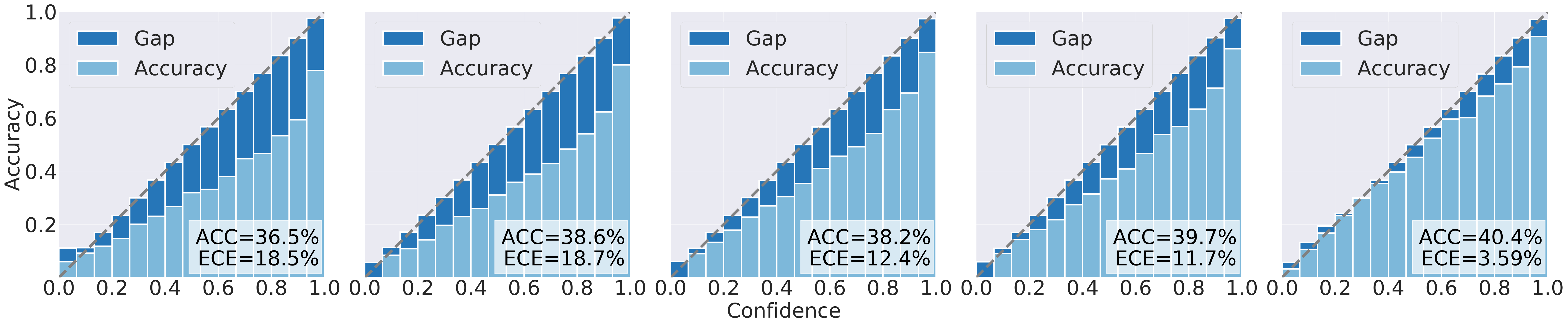} \\
		\vspace{-5pt}
		\caption{Reliability diagrams of ResNet-152 trained on Places-LT with 15 bins. From left to right: cRT, LWS, cRT with mixup, LWS with mixup, and MiSLAS.} 
		\vspace{-10pt}
		\label{fig:more_rd_place}
	\end{figure}
	
	\begin{figure}[h]
		\begin{subfigure}{0.195\linewidth}
			\begin{center}
				{\quad \quad cRT}
			\end{center}
		\end{subfigure}
		\hfill
		\begin{subfigure}{0.195\linewidth}
			\centering
			{\quad   \hspace{1.5pt}  LWS}
		\end{subfigure}
		\hfill
		\begin{subfigure}{0.195\linewidth}
			\centering
			{\quad {mixup + cRT}}
		\end{subfigure} 
		\hfill
		\begin{subfigure}{0.195\linewidth}
			\centering
			{\quad \hspace{-5pt}  mixup + LWS}
		\end{subfigure} 
		\hfill
		\begin{subfigure}{0.195\linewidth}
			\centering
			{MiSLAS}
		\end{subfigure}\\
		\centering
		\includegraphics[width=\textwidth]{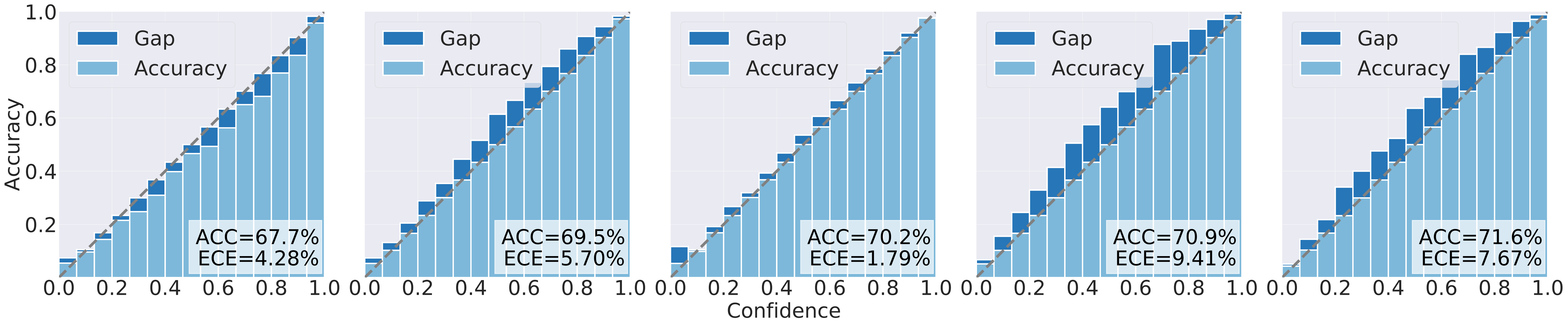} \\
		\vspace{-5pt}
		\caption{Reliability diagrams of ResNet-50 trained on iNaturalist 2018 with 15 bins. From left to right: cRT, LWS (under-confidence), cRT with mixup, LWS with mixup (under-confidence), and MiSLAS (under-confidence).} 
		\vspace{-10pt}
		\label{fig:more_rd_inat}
	\end{figure}
	
	\clearpage
	\section{More Results on ImageNet-LT, iNaturalist 2018, and Places-LT}\label{sec:more_result}
	\begin{table*}[h]
		\setlength{\tabcolsep}{20pt}
		\centering
		\begin{tabular}{l|l|cccc}
			\toprule[1.5pt]
			{\textbf{Backbone}}		& {\textbf{Method}}	 & Many & Medium & Few & All	\\ 
			\midrule
			{\multirow{5.1}*{ResNet-50}} & cRT & 62.5 & 47.4 & 29.5 & 50.3 \\
			& LWS & 61.8 & 48.6 & 33.5 & 51.2 \\
			& cRT+mixup & \textbf{63.9} & 49.1 & 30.2 & 51.7 \\
			& LWS+mixup& 62.9 & 49.8 & 31.6 & 52.0 \\
			& MiSLAS & 61.7 & \textbf{51.3} & \textbf{35.8} & \textbf{52.7} \\
			\midrule
			{\multirow{5.1}*{ResNet-101}} & cRT& 63.8 & 48.5 & 30.0 & 51.4 \\
			& LWS & 63.1 & 49.9 & 33.8 & 52.3 \\
			& cRT+mixup& \textbf{65.2} & 50.6 & 31.6 & 53.1 \\
			& LWS+mixup& 64.5 & 51.2 & 34.1 & 53.5 \\
			& MiSLAS& 64.3 & \textbf{52.1} & \textbf{35.8} & \textbf{54.1} \\
			\midrule
			{\multirow{5.1}*{ResNet-152}} & cRT	& 64.9 & 50.4 & 30.6 & 52.7 \\
			& LWS & 64.1 & 51.8 & 35.5 & 53.8 \\
			& cRT+mixup & \textbf{66.5} & 51.6 & 32.8 & 54.2 \\
			& LWS+mixup & 66.1 & 52.2 & 34.5 & 54.6 \\
			& MiSLAS & 65.4 & \textbf{53.2} & \textbf{37.1} & \textbf{55.2} \\
			\bottomrule[1.5pt]     
		\end{tabular}
		\caption{Comprehensive accuracy results on ImageNet-LT with different backbone networks (ResNet-50, ResNet-101 \& ResNet-152) and training 180 epochs.}
		\vspace{-5pt}
		\label{tab:imgnet_more}
	\end{table*}
	
	\begin{table*}[h]
		\setlength{\tabcolsep}{20.5pt}
		\centering
		\begin{tabular}{l|l|cccc}
			\toprule[1.5pt]
			{\textbf{Backbone}}		& {\textbf{Method}}	 & Many & Medium & Few & All	\\ 
			\midrule
			{\multirow{6.5}*{ResNet-50}} & cRT	& 73.2 & 68.8 & 66.1 & 68.2\\
			&$\tau$-normalized& 71.1 & 68.9 & 69.3 & 69.3\\
			&LWS& 71.0 & 69.8 & 68.8 & 69.5\\
			\cmidrule(lr){2-6}
			& cRT+mixup& \textbf{74.2}  & 71.1  &  68.2  & 70.2   \\
			& LWS+mixup& 72.8 & 71.6  & 69.8  & 70.9    \\
			& MiSLAS & 73.2  & \textbf{72.4}  &  \textbf{70.4} & \textbf{71.6}    \\
			\bottomrule[1.5pt]     
		\end{tabular}
		\caption{Comprehensive accuracy results on iNaturalist 2018 with ResNet-50 and training 200 epochs.}
		\vspace{-5pt}
		\label{tab:inua_more}
	\end{table*}

	\begin{table*}[h]
		\setlength{\tabcolsep}{20pt}
		\centering
		\begin{tabular}{l|l|cccc}
			\toprule[1.5pt]
			{\textbf{Backbone}}	& {\textbf{Method}}	 & Many & Medium & Few & All	\\ 
			\midrule
			{\multirow{13}*{ResNet-152}} &Lifted Loss& 41.1 &35.4 &24.0 & 35.2 \\
			&Focal Loss &41.1 &34.8 &22.4 &34.6 \\
			&Range Loss &41.1 &35.4 &23.2 &35.1 \\
			&FSLwF &43.9 &29.9& 29.5 &34.9 \\
			&OLTR &\textbf{44.7} &37.0 &25.3 &35.9  \\
			\cmidrule(lr){2-6}
			&OLTR+LFME &39.3 &39.6 &24.2 &36.2  \\
			&cRT &42.0 &37.6 &24.9 &36.7\\
			&$\tau$-normalized &37.8 &40.7 &31.8 &37.9\\
			&LWS &40.6 &39.1 &28.6 &37.6 \\
			\cmidrule(lr){2-6}
			&cRT+mixup& 44.1 & 38.5  &  27.1  & 38.1    \\
			&LWS+mixup& 41.7  & 41.3  &  33.1  & 39.7    \\
			&MiSLAS & 39.6 & \textbf{43.3} & \textbf{36.1}   & \textbf{40.4}    \\
			\bottomrule[1.5pt]     
		\end{tabular}
		\caption{Detailed accuracy results on Places-LT, starting from an ImageNet pre-trained ResNet-152.}
		\vspace{-5pt}
		\label{tab:place_more}
	\end{table*}
	\clearpage
	\section{Proof of Eq.~(\ref{eqn:solution_las}), the Optimal Solution of LAS}\label{sec:proof}
	In this section, we prove the optimal solutions of cross-entropy, the re-weighting method, and LAS. Furthermore, the comparison among above three methods will also be discussed.
	
	The general loss function form of these three methods for $K$ classes can be written as
	\begin{equation}\label{eqn:general_form}
	l= -\sum_{i=1}^{K}\vq_i\log\vp_i, \quad \quad \vp_i = \text{softmax}(\vw_i^\top \vx), \quad \quad  s.t., \quad \sum_i^K\vp_i=1, \tag{8}
	\end{equation}
	where $\vp$, $\vw$, and $\vx$ are the predicted probability, the weight parameter of the last fully-connected layer, and the input of the last fully-connected layer, respectively. 
	When the target label $\vq$ is defined as
	\begin{equation}
	\begin{split}
	\quad \vq_i=\left\{
	\begin{array}{ll}
	1, \quad & i= y,\\
	0, & i \neq y,\\
	\end{array}\right.
	\end{split}\notag
	\end{equation}
	where $y$ is the original ground truth label. Eq.~(\ref{eqn:general_form}) becomes the commonly used cross-entropy loss function. Similarly, when the target label $\vq$ is defined as
	\begin{equation}
	\begin{split}
	\quad \vq_i=\left\{
	\begin{array}{ll}
	w_i, \quad & i = y, \ \  \text{and} \ \  w_i > 0, \\
	0, & i \neq y, \\
	\end{array}\right.
	\end{split}\notag
	\end{equation}
	Eq.~(\ref{eqn:general_form}) becomes the re-weighting loss function. Moreover, 	when the target label $\vq$ is
	\begin{equation}
	\begin{split}
	\quad \vq_i=\left\{
	\begin{array}{ll}
	1-\epsilon_y = 1 - f(N_y), \quad & i= y,\\
	\frac{\epsilon_y}{K-1} = \frac{f(N_y)}{K-1}, &i \neq y,\\
	\end{array}\right.
	\end{split}\tag{9}
	\label{eqn:lass_q}
	\end{equation}
	Eq.~(\ref{eqn:general_form}) becomes the proposed LAS method. To get the optimal solution of Eq.~(\ref{eqn:general_form}), we define its Lagrange multiplier form as
	\begin{equation}\label{eqn:Lagrange_form}
	L = l + \lambda\left(\sum_i^K\vp_i -1\right) = -\sum_{i=1}^{K}\vq_i\log\vp_i + \lambda\left(\sum_i^K\vp_i -1\right), \tag{10}
	\end{equation}
	where $\lambda$ is the Lagrange multiplier. The first order conditions of Eq.~(\ref{eqn:Lagrange_form}) \textit{w.r.t.}  $\lambda$ and $\vp$ can be written as
	\begin{equation}
	\label{eqn:grad}
	\begin{aligned}
	\frac{\partial L}{\partial \lambda} &=& \sum_{i=1}^{K}\vp_i - 1 &=&0,  \\
	\frac{\partial L}{\partial \vp_i} &=& -\frac{\vq_i}{\vp_i} + \lambda &=&0. 
	\end{aligned} \tag{11}
	\end{equation}
	According to Eq.~(\ref{eqn:grad}), we get $\vp_i = \frac{\vq_i}{\sum_{j=1}^{K}\vq_j}$. Then, in the case of cross-entropy and re-weighting loss function, we get $\vp_i=1, i=y$ and $\vp_i=0, i \neq y$. Noting that 
	$$\vp_i = \text{softmax}(\vw_i^\top \vx)=\frac{\exp(\vw_i^\top \vx)}{\sum_{j=1}^{K}\exp(\vw_j^\top \vx)},$$
	the optimal solutions of $\vw_i^\top\vx$ for both cross-entropy and re-weighting loss functions are the same, that is, ${\vw_i^{*}}^\top\vx = \inf$. This means that both cross-entropy and re-weighting loss functions make the weight vector of the right class $\vw_i, i=y$ large enough while the others $\vw_j, j\neq y$ sufficiently small. As a result, they cannot change the predicted distribution and relieve over-confidence effectively. In contrast, in our LAS, according to Eqs.~(\ref{eqn:lass_q}) and (\ref{eqn:grad}), we get
	\begin{equation}
	\vp_i = \frac{\exp(\vw_i^\top \vx)}{\sum_{j=1}^{K}\exp(\vw_j^\top \vx)}  = \frac{\vq_i}{\sum_{j=1}^{K}\vq_j}=\left\{
	\begin{array}{ll}
	1-\epsilon_y,  \quad& i=y,\\
	\frac{\epsilon_y}{K-1},  \quad& i\neq y,\\
	\end{array}\right.
	\ \Longrightarrow\  \quad {\vw_i^*}^\top\vx=\left\{
	\begin{array}{ll}
	\log\left[\frac{(K-1)(1-\epsilon_y)}{\epsilon_y}\right] + c, \quad & i = y,\\
	c, \quad &i\neq y,\\
	\end{array}\right. \tag{12}
	\end{equation}
	where $c \in \sR$ can be an arbitrary real number. Overall, comparing with the infinite optimal solution in cross-entropy and re-weighting method, LAS encourages a finite output, which leads to a more general result, properly refines the predicted distributions of the head, medium, and tailed classes, and remedies over-confidence effectively.
	
	\clearpage
	
	\section{More Results about the Effect of mixup on cRT and LWS}
	
	\begin{figure*}[h]
		\centering
		\begin{subfigure}{0.48\linewidth}
			\includegraphics[width=\textwidth]{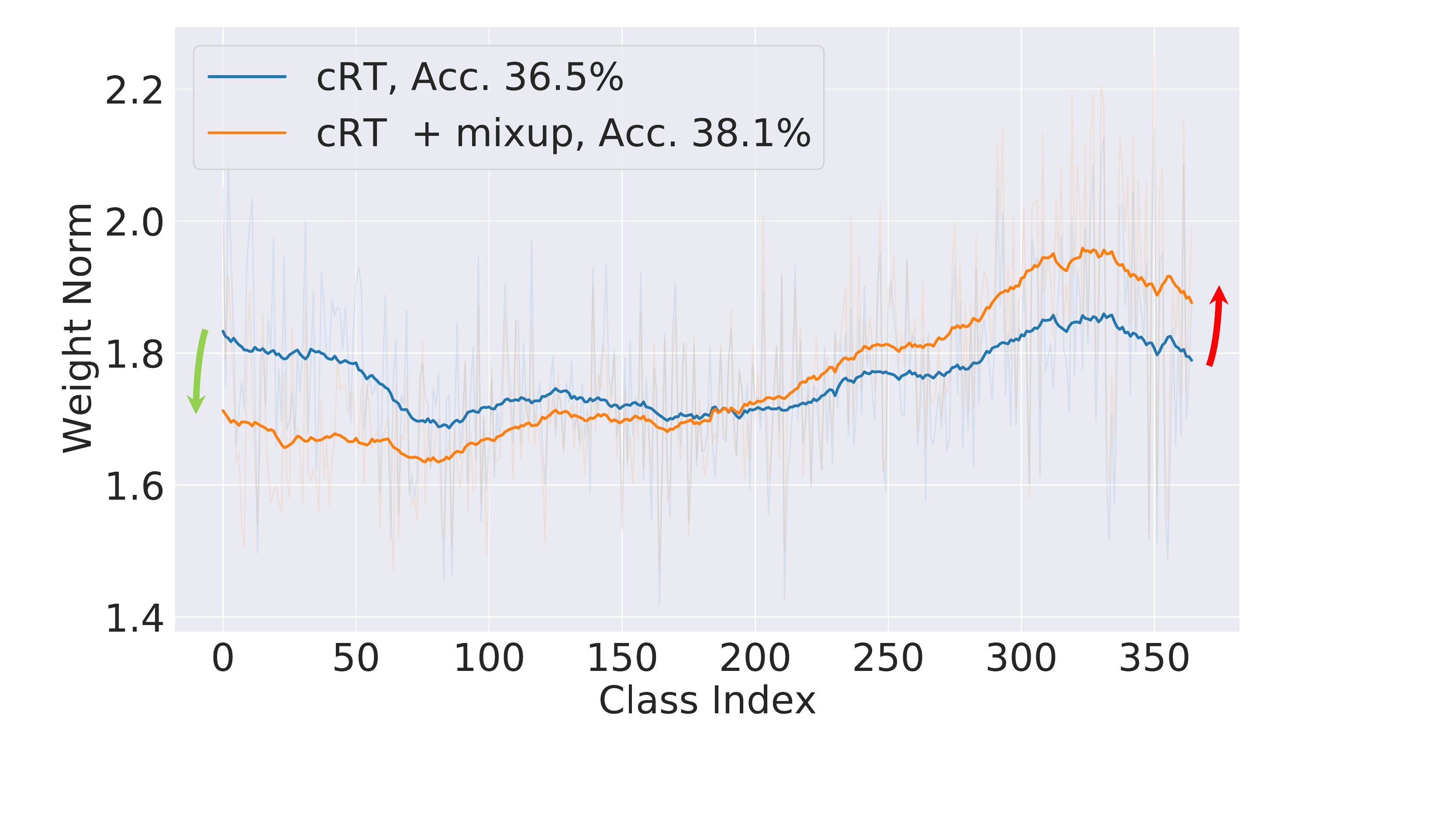}
		\end{subfigure}
		\hfill
		\begin{subfigure}{0.48\linewidth}
			\includegraphics[width=\textwidth]{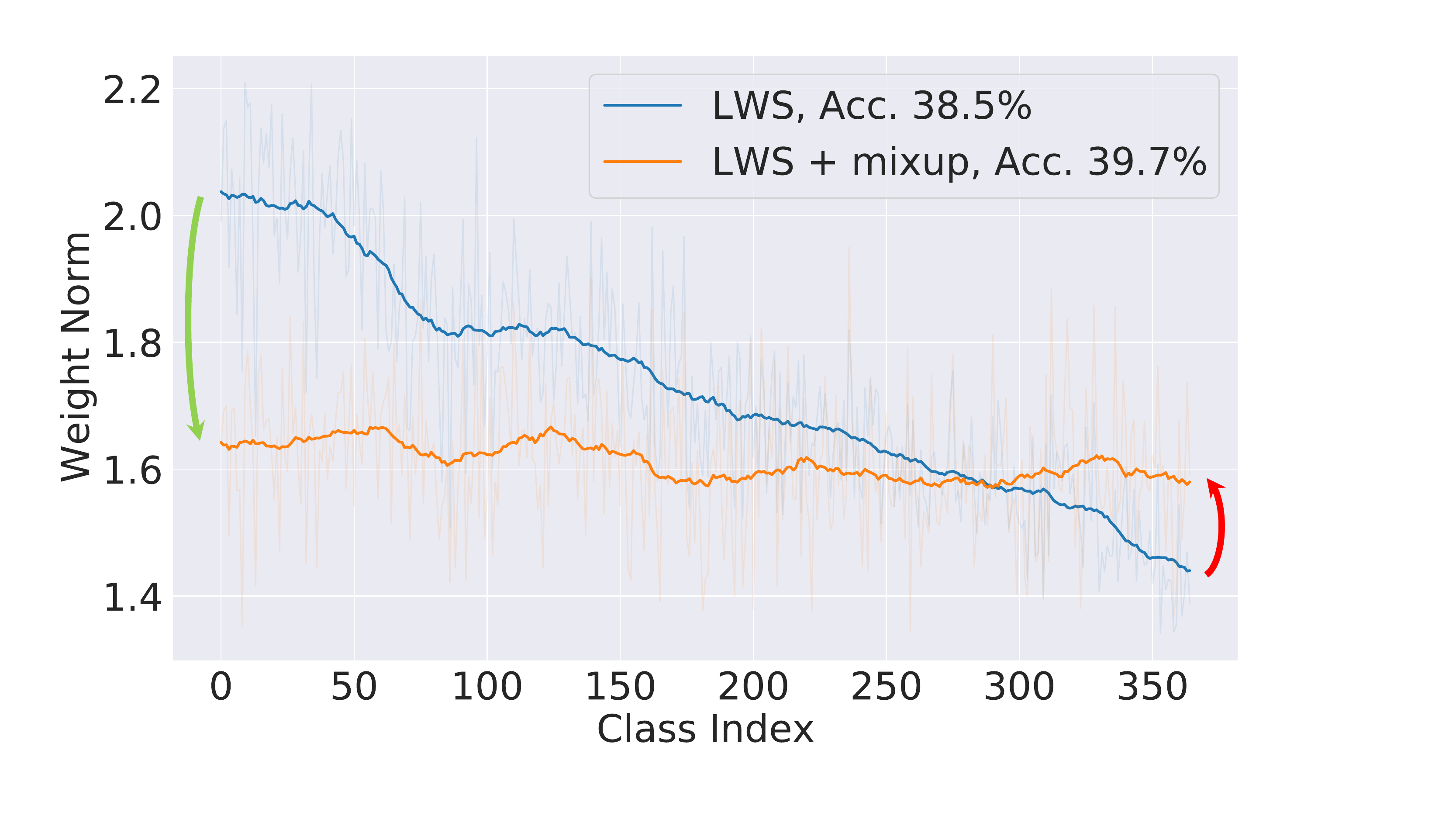}
		\end{subfigure}
		\vspace{-2pt}
		\caption{Classifier weight norms for the Places-LT evaluation set (365 classes in total) when classes are sorted by descending values of $N_j$, where $N_j$ denotes the number of training sample for
			Class-$j$. Left: weight norms of cRT with/without mixup. Right: weight norms of LWS with/without mixup. Light shade: true norm. Dark lines: smooth version. \textit{Best viewed on screen}.}
		\label{fig:norm1}
		
	\end{figure*}
	
	\begin{figure*}[h]
		\centering
		\begin{subfigure}{0.48\linewidth}
			\includegraphics[width=\textwidth]{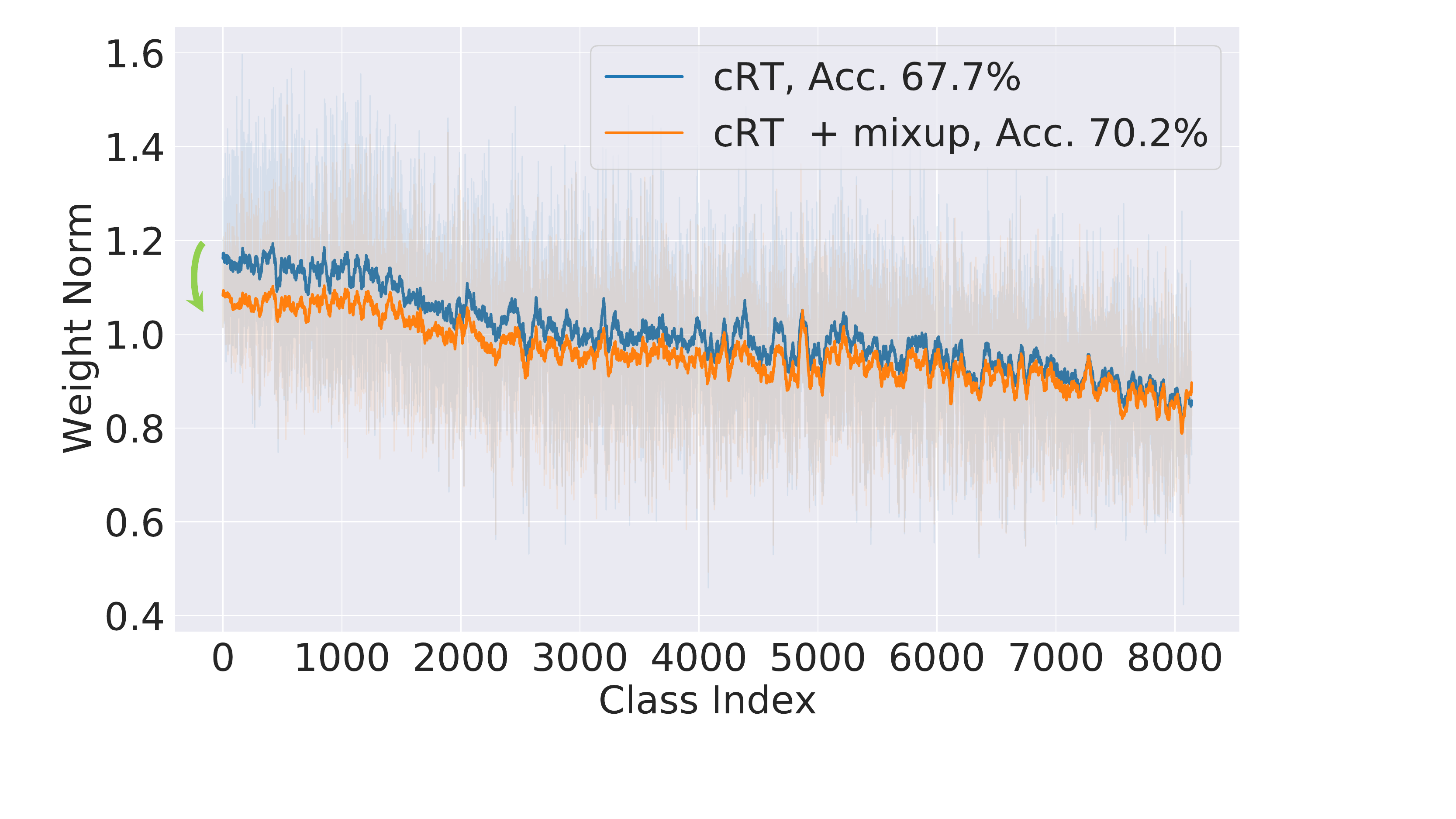}
		\end{subfigure}
		\hfill
		\begin{subfigure}{0.48\linewidth}
			\includegraphics[width=\textwidth]{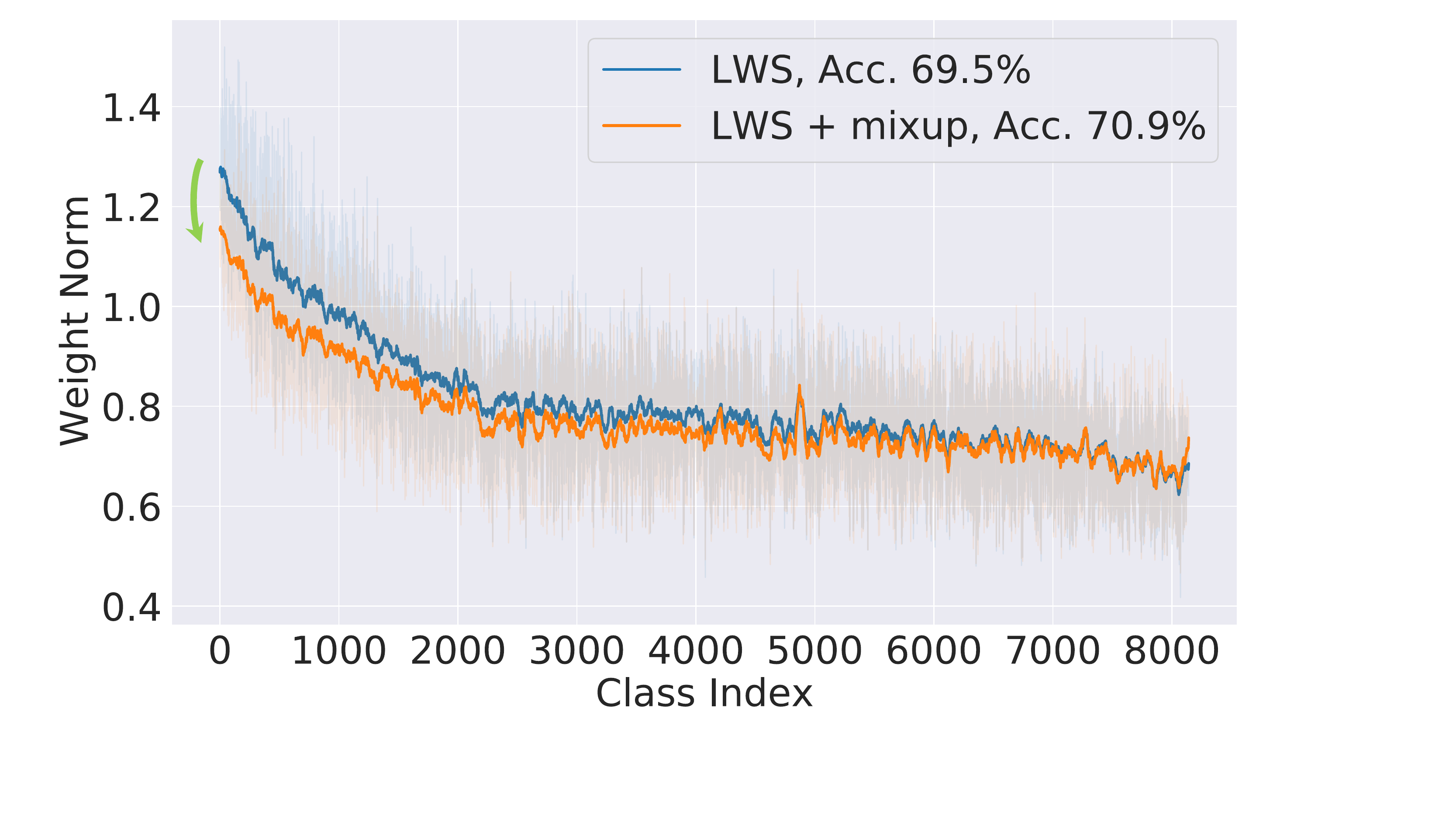}
		\end{subfigure}
		\caption{Classifier weight norms for the iNaturalist 2018 validation set (8,142 classes in total) when classes are sorted by descending values of $N_j$, where $N_j$ denotes the number of training sample for Class-$j$. Left: weight norms of cRT with or without mixup. Right: weight norms of LWS with or without mixup. Light shade: true norm. Dark lines: smooth version. \textit{Best viewed on screen}.}
		\label{fig:norm2}
	\end{figure*}
	
	As mentioned in Sec.~\ref{sec:mixup} and Fig.~\ref{fig:norm}, we observe that when applying mixup (orange line), the weight norms of the tail classes tend to be larger and the weight norms of the head classes are decreased, which means mixup may be more friendly to the tail classes. Here, we show more evidences that mixup reduces dominance of the head classes. In Figs.~\ref{fig:norm1} and \ref{fig:norm2}, norm of these variants are trained on Places-LT and iNaturalist 2018, respectively. The results are similar and consistent with those trained on ImageNet-LT.

\end{document}